\documentclass[final,9pt,times]{elsarticle}

\usepackage{booktabs}
\usepackage{hyperref}
\makeatletter
\def\ps@pprintTitle{%
 \let\@oddhead\@empty
 \let\@evenhead\@empty
 \def\@oddfoot{}%
 \let\@evenfoot\@oddfoot}
\makeatother

\usepackage{graphicx}
\usepackage{amssymb}
\usepackage{commath}
\usepackage{mathtools}
\usepackage{lineno}
\usepackage{multirow}
\usepackage{subcaption}
\usepackage{float}

\begin{document}

\begin{frontmatter}

\vspace*{\fill}
\begin{center}
Draft - work in progress. \\ 
Currently under initial review, mistakes may be present and content is likely to be revisited in subsequent versions.
\end{center}
\vspace*{\fill}
    
\title{SimPool: Towards Topology Based Graph Pooling with Structural Similarity Features}
\author{Yaniv Shulman}
\address{yaniv@aleph-zero.info}

\begin{abstract}
Deep learning methods for graphs have seen rapid progress in recent years with much focus awarded to generalising Convolutional Neural Networks (CNN) to graph data. CNNs are typically realised by alternating convolutional and pooling layers where the pooling layers subsample the grid and exchange spatial or temporal resolution for increased feature dimensionality. Whereas the generalised convolution operator for graphs has been studied extensively and proven useful, hierarchical coarsening of graphs is still challenging since nodes in graphs have no spatial locality and no natural order. This paper proposes two main contributions, the first is a differential module calculating \emph{structural similarity features} based on the adjacency matrix. These structural similarity features may be used with various algorithms however in this paper the focus and the second main contribution is on integrating these features with a revisited pooling layer DiffPool \cite{NIPS2018_7729} to propose a pooling layer referred to as \emph{SimPool}.This is achieved by linking the concept of network reduction by means of structural similarity in graphs with the concept of hierarchical localised pooling. Experimental results demonstrate that as part of an end-to-end Graph Neural Network architecture SimPool calculates node pooling assignments that functionally resemble more to the locality preserving pooling operations used by CNNs that operate on local receptive fields in the standard grid. Furthermore the experimental results demonstrate that these features are useful in inductive graph classification tasks with no increase to the number of parameters.
\end{abstract}
\end{frontmatter}


\section{Introduction}
\label{s:Introduction}
Deep learning methods have proven very successful at capturing hidden patterns of Euclidean data and obtained state-of-the-art results in various applications such as time series regression, similarity metric learning, machine vision and natural language processing. However there are many applications where data is best represented in the form of graphs such as the representation of molecules in chemistry, relationships networks, recommender systems and applications in traffic management \cite{Wu_2020}. Machine learning algorithms that are designed to operate on regular grid data are not readily applicable to graphs since these generally have irregular structure with varying sizes of unordered nodes \cite{hamilton2017representation}. Due to the irregularity of graphs and lack of natural order of nodes and edges it is challenging to generalize some common grid operations such as shifting, convolutions and coarsening to arbitrary graphs. However in recent years much progress has been made towards unifying deep learning frameworks that operate of regular grids and learning frameworks for graphs of arbitrary topology \citep{7974879}. In particular one such family of models are Graph Neural Networks (GNN) that utilise similar mechanics to other Artificial Nueral Networks (ANN) and Convolutional Neural Networks (CNN) \cite{Goodfellow-et-al-2016} to enable a unified approach to learning on both standard grid and arbitrary graph represented data \cite{hamilton2017representation}. Similarly to CNNs, GNNs learn representations of nodes and graphs as points in a vector space $\mathbb{R}^d$ so that geometric relationships between vectors in the embedding space provide sufficient information about the nodes and graphs structure to solve a learning task. An important aspect of end-to-end CNNs operating on regular grid input is the hierarchical coarsening of the grid by localised pooling. CNNs take advantage of the stationarity (shift-invariance) and  compositionality of grid arranged data by utilsing local statistics and impose a prior on the data by virtue of the CNN architecture \cite{7974879}. CNNs are typically realised by alternating convolutional and pooling layers where the pooling layers subsample the grid and exchange spatial or temporal resolution for increased feature dimensionality. The output features are translation invariance/covariance depending on whether hierarchical grid coarsening is performed by means of pooling or kept constant \cite{7974879}. 

Whereas pooling operations can be naturally defined on generalized grid graphs, extending these operations to graph data of arbitrary topology is challenging since nodes in graphs have no spatial locality and no natural order. However coarsening of the graph is a critical step for generating graph embeddings and is required as a minimum at least once at the point of merging the individual node embeddings to a representation of the entire graph such as the models in \cite{DBLP:conf/icml/LiGDVK19, 7974879, Zhang2018AnED, 10.5555/3305381.3305512}. Transformation of all nodes into a single representation is referred to as global pooling and has usefulness in inductive learning tasks where graphs typically have different number of nodes. However global pooling does not take the topology of the graph into account since all node embeddings are aggregated at once using the same procedure. Whereas much attention was put into GNNs in general including global pooling strategies, the hierarchical coarsening of graphs as part of an end-to-end graph embedding network does not seem to be as intensely researched. 

In social studies the concept of \emph{structural equivalence} \cite{lorrain1971structural} is an important explanotory factor in the study of social homogenity \cite{Borgatti_Structural_Equivalence}. A common definition of structural equivalence is that two nodes are structurally equivalent if they share the same neighbourhoods. Given an undirected graph $G(V,E)$, a pair of nodes $i,j \in V$ are structurally equivalent if $N(i) = N(j)$, where $N(\cdot)$ is the set of neighbouring nodes (connected with an edge) \cite{leicht2005vertex}. For directed graphs, the definition changes such that $i$ and $j$ have incoming edges from the same nodes and outgoing edges to the same nodes. Structural equivalence was introduced as a method for reducing models of social networks e.g. to block models. However it rapidly gained importance as an approach to formalise the concept of relational role or position based on the concept that structurally equivalent nodes share many social attributes \cite{Borgatti_Structural_Equivalence}.

In the remainder of this paper the focus is on bridging the gap between CNNs and GNNs to enable GNNs to perform hierarchical coarsening of a graph similarly to CNNs that operate on local receptive fields in the standard grid such that nearby nodes are more likely to belong to the same cluster in the coarsened graph. This is achieved by linking the concept of network reduction by means of structural similarity in graphs with the concept of hierarchical localised pooling. To summarise the key contributions of this work are:

\begin{enumerate}
\item A differential module for calculating \emph{structural similarity features} based on the adjacency matrix by defining a differential representation of the top-$k$ $argmax$ operator that is applicable to graphs of various sizes. Furthermore these features may be used in conjunction with various algorithms to calculate node pooling assignments or used in learning tasks where graph structure conveys critical information and augment or replace the node features.
\item Revisiting the DiffPool algorithm proposed in \cite{NIPS2018_7729} and integrating with the structural similarity features to propose SimPool, a pooling layer which calculates node pooling assignments that functionally resemble more to the locality preserving pooling operations used by CNNs that operate on local receptive fields in the standard grid.
\end{enumerate}

\section{Related Work}
\label{s:Related Work}
There has been extensive research on GNNs in recent years \cite{Wu_2020, hamilton2017representation}. These include methods related to spectral graph convolutions that generalise a convolutional network through the Graph Fourier Transform \cite{DBLP:journals/corr/BrunaZSL13, Henaff2015DeepCN,10.5555/3157382.3157527, DBLP:journals/corr/KipfW16}. GNNs such as \cite{DBLP:journals/corr/KipfW16, column_networks, graphsage, DBLP:conf/icml/LiGDVK19, simonovsky2017dynamic} typically use an approach of generating embeddings for a node or a graph by iteratively aggregating the features of neighbouring nodes. These methods feature a number of desirable attributes such as localised representations, incorporating graph structure, leverage node features and can be used in inductive learning settings as they are capable of generating embeddings for nodes or graphs not present during training \cite{hamilton2017representation}.
 
Previous research includes hierarchical coarsening of graphs by combining GNNs with deterministic graph clustering algorithms such as the SortPooling layer proposed in \cite{Zhang2018AnED} that sorts the node features consistently before inputting them into a 1-D convolutional and dense layers, the use of the VoxelGrid algorithm in \cite{simonovsky2017dynamic} and the use of Graclus algorithm \cite{10.1109/TPAMI.2007.1115} for clustering combined with localised spectral convolution \cite{10.5555/3157382.3157527}. A somewhat related pooling approach is suggested in \cite{DBLP:conf/icml/GaoJ19} that performes down-sampling on graph data by selecting the top-$k$ subset of nodes having the largest projection magnitude on a 1-D trainable projection vector. A different end-to-end approach is taken by \cite{NIPS2018_7729} where in each pooling layer a differentiable soft assignment matrix is learned by a GNN which computes assignment weights for every node to one of the clusters in the coarsened graph. A self-attention based method coined SAGPool is proposed in \cite{35137c0f4e904fc0ab786021ead07852} where an attention mask is computed by a GNN to determine which nodes are selected to be passed on to the next layer. 

\section{Problem Description}
Methods utilising node features for calculating cluster assignments or node selection will by design assign nodes having similar neighbourhood features to the same cluster. Methods that select nodes by sorting, ranking or  an attention mechanism where such operations are based on node features will also select by design nodes having similar features and neighbourhoods thus lose information simply due to weak role similarity with some other real or virtual nodes. These issues are common to all the methods that perform coarsening by message passing GNNs such as \cite{Zhang2018AnED, DBLP:conf/icml/GaoJ19, NIPS2018_7729, 35137c0f4e904fc0ab786021ead07852}. This type of clustering/filtering is strongly related to the notion of role similarity and is likely, especially in graphs with repeating structures (e.g. molecular datasets), to assign distant nodes to the same cluster \cite{NIPS2018_7729}. Whilst role similarity based clustering is a useful concept in its own right it is dissimilar to the pooling layers of grid graphs typically used by CNNs that compose receptive fields of adjacent non-overlapping partitions of the data and thus are able to leverage local statistics of the data such as stationarity and compositionality \cite{7974879}.

\section{Preliminaries}
\label{S:Preliminaries}
In this section a summary of related methods is given to set the notation and naming conventions used in subsequent sections. Let $G = (V,E)$ denote a graph where $V$ and $E$ are sets of nodes and edges respectively. Typically each node $i \in V$ is associated with a node feature vector $\mathbf{x}_i \in \mathbb{R}^{d_0}$, and each edge $(i, j) \in E$ is associated with an edge feature vector $\mathbf{x}_{ij}$. Furthermore let $|V|$ denote the cardinality of $V$ (i.e., number of elements in $V$); $n_l$ denote the number of nodes (clusters) and $d_l$ the dimensionality of the node feature vectors $\{ \mathbf{x}_{i}\}_{i \in V}$ in layer $l$ respectively.

\subsection{Graph matching networks}
Graph Matching Networks (GMN) \cite{DBLP:conf/icml/LiGDVK19} propose a message passing global pooling GNN architecture that transformes a set of node and edge features into an embedding vector. GMN introduce three layers, two of which, an \emph{encoder} layer and the \emph{propogation} layer are used in this work. The encoder layer defined by \cite{DBLP:conf/icml/LiGDVK19} transforms separately the node and edge features as follows:
\begin{align}
\label{eq:gmn_encoder}
\begin{split}
\mathbf{h}_{i}^{(0)}&=MLP_{node}(\mathbf{x}_i), \quad \forall i \in V \\
\mathbf{e}_{ij} &= MLP_{edge}(\mathbf{x}_{ij}), \quad \forall (i, j) \in E
\end{split}
\end{align}

The propagation layer transforms a set of node representations $\{\mathbf{h}_i^{(t)}\}_{i \in V}$ to new node representations $\{\mathbf{h}_i^{(t+1)}\}_{i \in V}$ as follows:

\begin{align}
\label{eq:gmn_propogation}
\begin{split}
\mathbf{m}_{j \rightarrow i }&=f_{message}(\mathbf{h}_i^{(t)}, \mathbf{h}_j^{(t)}, \mathbf{e}_{ij}) \\
\mathbf{h}_{i}^{(t+1)}&=f_{node} \left( \mathbf{h}_{i}^{(t)}, \sum\nolimits_{j:(j,i) \in E} \mathbf{m}_{j \rightarrow i} \right)
\end{split}
\end{align}

Where $f_{message}$ is usually an MLP (multi-layer perceptron) on the concatenated inputs, and $f_{node}$ can be either an MLP or a Recurrent Neural Network (RNN) core. The encoder is normally the first hidden layer in the GMN model and by stacking multiple layers of the propagation layer the representation for each node will accumulate information from an increasing neighbourhood size.

\subsection{Graph convolutional network}
One commonly used variant of approximate spectral convultion GNN is the Graph Convolutional Network (GCN) \cite{DBLP:journals/corr/KipfW16} having the layer forward propagation rule:

\begin{align}
\label{eq:gcn_propogation}
\begin{split}
H^{(l+1)} & = \sigma \left( \tilde{D}^{-\frac{1}{2}} \tilde{A} \tilde{D}^{-\frac{1}{2}}H^{(l)}W^{(l)} \right) \\
\tilde{D}_{ii} &= \sum_j \tilde{A}_{ij} \\
\tilde{A} &= A + I_{|V|}
\end{split}
\end{align}

Where $\tilde{A}_{|V| \times |V|}$ is the undirected adjacency matrix of the graph with added self-connections; $I_{|V|}$ is the identity matrix; $W^{(l)}$ is the trainable weight matrix for layer $l$; $\sigma(\cdot)$ is an activation function; $H^{(l)} \in \mathbb{R}^{|V| \times d_l}$ is the matrix of activations in the $l$-th layer; $H^{(0)} = X$; and $X \in \mathbb{R}^{|V| \times d_0}$ is the nodes feature matrix.

\subsection{DiffPool}
DiffPool \cite{NIPS2018_7729} enables the construction of deep end-to-end multi-layer GNN models with hierarchic pooling by incorporating a differentiable layer that pools graph nodes. At the core of the model is an assignment matrix $S_{n_l \times n_{l+1}}^{(l)}$ calculated by a GNN that learns soft cluster assignments of each node at pooling layer $l$ to a cluster in the following coarsened pooling layer $l + 1$. Where each $row_i(S^{(l)})$ denotes the probabilities of node $i$ in pooling layer $l$ to be assigned to each of the $n_{l+1}$ clusters in pooling layer $l+1$. 

\begin{align}
\label{eq:diffpool}
\begin{split}
X^{(l+1)} &= S^{(l)^T} Z^{(l)} \in \mathbb{R}^{n_{l+1} \times d_{l + 1}} \\
A^{(l+1)} &= S^{(l)^T} A^{(l)} S^{(l)} \in \mathbb{R}^{n_{l+1} \times n_{l+1}} \\
Z^{(l)} &= GNN_{embed}^{(l)} (A^{(l)}, X^{(l)}) \in \mathbb{R}^{n_l \times d_{l+1}} \\
S^{(l)} &= softmax \left( GNN_{pool}^{(l)} (A^{(l)}, X^{(l)}) \right)
\end{split}
\end{align}

Where $A_{n_l \times n_l}^{(l)}$ and $X^{(l)} \in \mathbb{R}^{n_l \times d_l}$ are the graph adjacency matrix and the nodes feature matrix in layer $l$ respectively.

\section{SimPool}
\label{S:SimPool}
\subsection{Graph structural similarity matrix}
\label{S:Graph structural similarity matrix}
Let $C \in \mathbb{R}^{|V| \times |V|}, \quad C_{i,j} \coloneqq c(i,j), \quad i,j \in V$ be a graph similarity matrix where $c(\cdot, \cdot)$ is a real-valued function that quantifies the similarity between two nodes in $V$. In particular let $C^{cos,p}$ denote a graph structural similarity matrix where its calculation is defined as:

\begin{align}
\label{eq:Symmetric C}
\begin{split}
& \hat{A} = (A + \lambda I_{|V|})^p, \quad p \in \mathbb{N}, \quad \lambda \in  \mathbb{R}_{\geq 0} \\
& c(\hat{\mathbf{a}}_i, \hat{\mathbf{a}}_j) \coloneqq \mathbb{R}^{|V|} \times \mathbb{R}^{|V|} \rightarrow \mathbb{R} = \frac{\hat{\mathbf{a}}_i\cdot \hat{\mathbf{a}}_j}{\norm{\hat{\mathbf{a}}_i} \norm{\hat{\mathbf{a}}_j}} \\
& C_{i,j}^{cos,p} = c(\hat{\mathbf{a}}_i, \hat{\mathbf{a}}_j)
\end{split}
\end{align}

Where $A$ is a symmetric adjacency matrix with optionally added self connections; $c(\hat{\mathbf{a}}_i, \hat{\mathbf{a}}_j)$ is the standard cosine similarity measure over $\mathbb{R}^{|V|}$; and $\hat{\mathbf{a}}_i, \hat{\mathbf{a}}_j$ are the $i$-th and $j$-th column vectors in $\hat{A}$ respectively. Adding self connections is important to improve similarity representation in certain instances such as when a graph is undirected and non-reflexive (i.e., no edges from a node to itself). In such graphs if $i$ and $j$ are connected, it is not possible for them to be structurally equivalent. In particular the similarity is zero for nodes that are connected directly by an edge but do not share any other common neighbours. \newline

In the case where $A$ is asymmetric the calculation of $C^{cos,p}$ is modified as such:
\begin{align}
\label{eq:Asymmetric C}
\begin{split}
& \hat{A} = (A + \lambda I_{|V|})^p, \quad p \in \mathbb{N}, \quad \lambda \in  \mathbb{R}_{\geq 0} \\
& \tilde{A} = concat(\hat{A}, \hat{A}^T) \\
& c(\tilde{\mathbf{a}}_i, \tilde{\mathbf{a}}_j) \coloneqq \mathbb{R}^{2|V|} \times \mathbb{R}^{2|V|} \rightarrow \mathbb{R} = \frac{row_i(\tilde{A}) \cdot row_j(\tilde{A})}{\norm{row_i(\tilde{A})} \norm{row_j(\tilde{A})}} \\
& C_{i,j}^{cos,p} \coloneqq c(\tilde{\mathbf{a}}_i, \tilde{\mathbf{a}}_j)
\end{split}
\end{align}

Where $concat$ concatenates its inputs along the columns returning a $|V| \times 2|V|$ matrix. Note that the choice of similarity measure $c(\cdot,\cdot)$ is flexible and other similarity (distance) measures can be used such as Hamming or $L_2$, however at this work the scope is limited to the cosine similarity measure as described in this section. \newline

\subsection{Structural similarity features}
\label{S:Structural similarity features}
The utility of $C^{cos,p}$ is apparent when a node $i \in V$ is associated with $row_i(C^{cos^p})$ where now the set $\{row_i(C^{cos,p})\}_{i \in V}$ is referred to as \emph{structural similarity features}. The size of the neighbourhood affecting the calculation of the similarity measure $c(\cdot, \cdot)$ is dependent upon and is controlled by $p$. Note that the parameter $p$ can be adjusted for each pooling layer independently which may be useful to increase in layers where connectivity is sparse. These features capture the topological similarity between nodes in the graph and can be utilised by various algorithms and optionally be combined with additional node features such as the node labels or features to calculate node cluster assignments that strongly relate to structural similarity. By utilising $\{row_i(C^{cos,p})\}_{i \in V}$ for pooling calculations such as node cluster assignments GNNs may now be more closely aligned with CNNs that operate on grids where typically the pooling operation depends only on structure, and role similarity is propagated through composition of filters that operate on the data.

\subsection{Indices mapping trick}
\label{S:Indices mapping trick}
In datasets that contain graphs of various sizes, as is typically the case in inductive learning tasks such as graph classification, the dimensionality of the structural similarity features varies from an example to example. Thus it is impossible to use these features with many standard ANN layers that require a fixed input dimensionality. Consequently additional processing of the structural similarity features is proposed that results in features of constant dimensionality. Intuitively it is vital to retain the indices of the nodes that are most similar to a given node and some notion of their significance. Unfortunately the standard $argmax$ operation for the top-$k$ similar nodes is not a differentiable operation and can not be trivially used with gradient based training. However a method referred to as the \emph{indices mapping trick} is proposed as an alternative differential representation of the top-$k$ $argmax$ operator:

\begin{align}
\label{eq:Indices Mapping Trick}
\begin{split}
\hat{C} &= \frac{(\alpha C^{cos,p} + \mathbf{1} \mathbf{n}^T) \odot notzero(C^{cos,p})}{\abs{V} + 1} \\
idx &= rank_{cols}(C^{cos,p},k) \\
\tilde{C} &= \hat{C}[idx]
\end{split}
\end{align}

Where $rank_{cols}$ returns the matrix $idx_{|V| \times k}$ containing the indices (assuming one-based indexing) of the top-$k$ values for each row of its inputs ordered in descending input value (not descending indices). These indices correspond to the top-$k$ most similar nodes for each node in order of descending similarity; $\alpha \in [0,1]$ is a scaler that determines the magnitude of separation between mapped indices where lower values of $\alpha$ increase separation but reduce the possibility of preserving information about similarity magnitude; $\mathbf{1}$ is a column vector of $1$ with dimension $|V|$; $\mathbf{n}^T = \{1,2,\cdots, \, \abs{V}\}$; $\odot$ is the Hadamard product operator; $notzero$ is an element-wise indicator operator that returns an $|V| \times k$ binary matrix where an output element is $1$ when the corresponding input element is different than $0$, and $0$ when the corresponding input element is $0$; and $\tilde{C}_{|V| \times k}$ is the final processed structural similarity features matrix. Note that if $k < |V|$ then $\tilde{C}$ is padded with zeroes.

Assuming a non-negative adjacency matrix, the resulting $\tilde{C}$ is a non-overlapping mapping of the graph node indices to the range $[0, 1 - \frac{1- \alpha}{|V| + 1}]$, organised in a matrix such that the first column of $\tilde{C}$ contains the mapped indices of the most similar node to each corresponding node, the second column denotes the mapped indices of the second most similar nodes as so forth. The indices mapping trick \eqref{eq:Indices Mapping Trick} fulfil the following three key criteria:
\begin{enumerate}
\item The typical sparseness of a $C^{cos,p}$ is exploited for dimensionality reduction in a way that retains the significant topological information inherent to the graph structural similarity features.
\item The dimensionality of the resulting features is constant to enable use of ANN layers that can accept only inputs of fixed dimensionality.
\item End-to-end differentiability is maintained to enable use of gradient based optimisation methods.
\end{enumerate}

When $k \ll |V|$ there is substantial redundancy in the calculation as defined in \eqref{eq:Indices Mapping Trick} therefore an alternative efficient method of calculation is suggested: 

\begin{align}
\label{eq:Indices Mapping Trick Efficient}
\begin{split}
& idx = rank_{cols}(C^{cos,p},k) \\
& \hat{C} = \alpha C^{cos,p}[idx] \\
& \tilde{C} = \frac{\hat{C} + cols(idx)}{|V| + 1} \\
& \tilde{C}[zeros(\hat{C})] = 0
\end{split}
\end{align}

Where $cols$ returns a  $idx_{|V| \times k}$ matrix containing the column indices only and $zeros$ returns the indices of zero valued elements in its matrix argument.

\subsection{Assignment matrix}
\label{S:Assignment matrix}
Following the definition in \cite{NIPS2018_7729} let $S_{n_l \times n_{l+1}}^{(l)}$ denote the learned soft cluster assignment matrix of each node at pooling layer $l$ to a cluster in the following coarsened pooling layer $l + 1$. To perform effective pooling, in each pooling layer an assignment matrix needs to be learned that considers the coarsened graph connectivity at that layer. Therefore it is desirable to use features related to the graph connectivity such as the structural similarity features to calculate the cluster assignments. To formulate the calculation of $S$ the isomorphism of graphs by means of permutation of the graph adjacency matrix need also be considered. First a proof that a permutation of the adjacency matrix results in $C^{cos,p}$ permuted in the same manner. \newline

\textbf{Proposition 1}. \emph{An isomorphic permutation of a symmetric adjacency matrix} $A$ \emph{results in} $C^{cos,p}$ \emph{permuted in the same manner. Formally, let} $\hat{A} = (A + \lambda I_{|V|})^p \in \mathbb{R}^{n_l \times n_l}, \: p \in \mathbb{N}, \: \lambda \in \mathbb{R}_{\geq 0}$ \emph{be a symmetric matrix, assume that} $\norm{\mathbf{\hat{a}_i}} = 1$ \emph{and let} $P \in \{0,1\}^{n_l \times n_l} = p_1 p_2 \cdots p_k$ \emph{be any permutation matrix where} $p_i$ \emph{is a symmetric elementary permutation matrix, then} $C^{cos,p} = \hat{A} \hat{A}^T \Rightarrow PC^{cos,p}P^T = P \hat{A}P^T(P \hat{A} P^T)^T$ \newline

\emph{Proof}: First establish that $P^T \hat{A} P = P \hat{A} P^T$. By symmetry and transposition identities $p^T \hat{A} p = p \hat{A}^Tp^T = (p^T \hat{A} p)^T$ therefore by reapplying the same reasoning repeatedly calculating the central brackets, $P^T \hat{A} P = p_k^T ( \cdots ( p_2^{T} ( p_{1}^{T} \hat{A} p_{1} ) p_2 ) \cdots ) p_k = (P^T \hat{A} P)^T = P \hat{A}^TP^T = P\hat{A} P^T$ concluding the proof that $P^T \hat{A} P = P \hat{A} P^T$. Furthermore by definition in \eqref{eq:Asymmetric C} $C^{cos,p} = \hat{A} \hat{A}^T$, therefore $PC^{cos,p}P^T = P \hat{A} \hat{A}^TP^T = P \hat{A} P^{-1}P \hat{A}^TP^T = P \hat{A} P^TP \hat{A}^TP^T = P \hat{A} P^T(P^T \hat{A} P)^T = P \hat{A}P^T(P \hat{A} P^T)^T$, concluding the proof of Proposition 1.\newline

Therefore using $C^{cos,p}$ for calculating cluster assignments is in the case of a symmetric $A$ is permutation invariant as long as the function used is permutation invariant (e.g. GMN). However permutations of the adjacency matrix in conjunction with the indices mapping trick \eqref{eq:Indices Mapping Trick} do not result in trivial permutations of $\tilde{C}$ but rather in a similar permutation of the rows of $\tilde{C}$ and in addition a consistent replacement of values in $\tilde{C}$ that reflect the permutation of indices in the graph.

When the graph sizes in the dataset are relatively small a RNN core can be used as the first stage in the calculation of $S^{(0)}$ since it can accept inputs of various lengths and calculate fixed sized outputs. The use of a RNN core can be stand-alone, combined with other subsequent dense layers or as the first component of a permutation invariant GNN layer such as the GMN encoder \cite{DBLP:conf/icml/LiGDVK19}. Utilising a RNN core is done without any loss of information. However when the graphs increase in size the use of a RNN can be prohibitive in terms of the computational resources and performance may be suboptimal due to the inherent difficulties in processing very long input sequences. Therefore the indices mapping trick \eqref{eq:Indices Mapping Trick} can be used to fix the dimensionality of the structural similarity features so that any standard ANN layer can be used.

\subsection{Complexity}
\label{S:Complexity}
When the indices mapping trick is used the cosine similarity calculation can be done iteratively on a subset of nodes and by retaining the indices and values of the top-$k$ most similar nodes only the storage complexity can be reduced to $\mathcal{O}(|V|)$. Furthermore since the similarity features are deterministically calculated from the adjacency matrix with no learned parameters involved, the calculation of the structural similarity features for the input graphs (where the majority of complexity lies) can be done offline once as a dataset preprocessing step. In addition observe that since $C_{ij}^{cos,p} = 0$ for any two nodes $i,j \in V$ having a geodesic distance larger than two. Therefore for many if not most real-world graphs a substantial reduction in computation can be achieved by calculating only the similarity between nodes having a geodesic distance of two or less. Consequently assuming a mean degree of $d_{\mu}$ the amortised complexity is $(d_{\mu} + d_{\mu}^2)|V| = \mathcal{O}(d_{\mu}^2|V|)$.

\subsection{DiffPool revised}
\label{S:DiffPool revised}
Having defined the structural similarity features they can now be utilised in an end-to-end heirearchical graph pooling architecture. For this purpose the DiffPool model suggested by \cite{NIPS2018_7729} is revised and the utility of the suggested structural similarity features as well as additional improvements to the original DiffPool algorithm are evaluated. Changes to the original DiffPool include the following:
\begin{enumerate}
\item A change to the calculation of $A^{(l+1)}$ the adjacency matrix at pooling layer $l+1$.
\item Changes to the calculation of $S^{(l)}$ the assignment matrix at pooling layer $l$.
\item Proposition of a new regularisation term to encourage the model to assign nodes to available clusters and to distribute nodes evenly across assigned clusters.
\item Removal of the auxiliary link prediction objective. \newline
\end{enumerate}

\textbf{Calculation of the adjacency matrix}. Typically all elements of $A^{(l+1)}$ pre-activation are non-negative therefore adding a $tanh$ activation restricts $A^{(l)}$ to the range $[0,1)$ post-activation. This change seemed to result in improved performance and increased stability during training.

\begin{equation}
\label{eq:New A_l+1}
A^{(l+1)} = tanh \left( S^{(l)^T} A^{(l)} S{(l)} \right)
\end{equation}

\textbf{Calculation of the assignment matrix}. Since the structural similarity features $\{row_i(C^{cos,p})\}_{i \in V}$ are calculated from the adjacency matrix and encode neighbourhood connectivity and in addition a similarity based order is imposed on the features when the indices mapping trick is applied there is no longer a necessity to use a GNN layer. On the contrary, using a MLP or a RNN can lead to consideration of information from all nodes when calculating cluster assignments rather than considering only the local neighbourhood of a node and may result in improved performance especially in deeper layers where the connectivity is high. Therefore it is suggested that $S^{(l)}$ maybe calculated with any arbitrary ANN layer or subnetwork, including GNNs. \newline

\textbf{Additional regularisation}. In the original DiffPool model \cite{NIPS2018_7729} the authors suggest to regularise cluster assignment by minimising $L_E$:

\begin{align}
\label{eq:L_E}
\begin{split}
L_E &= \sum_{l}L_E^{(l)} \\
L_E^{(l)} &= \frac{1}{n_l} \sum_{i=1}^{n_l} H(row_i(S^{(l)}))
\end{split}
\end{align}

Where $H(\cdot)$ denotes the entropy function. This regularisation term encourages the rows of $S$ to resemble one-hot-encoded vectors and therefore results in assignment of nodes that is close to "unique". However it was observed in experiments that this regularisation technique also encourages the model into effectively utilising a small number of clusters, typically as low as one or two clusters at most. This may explain the statement in \cite{NIPS2018_7729} that training the pooling GNN using only gradient signal wasn't effective and that training often converge to a spurious local minima early in training. To mitigate this behaviour it is proposed to incorporate an additional regularisation term. Intuitively there are two goals that are desirable to achieve, the first is encouraging the model to utilise as many clusters as it is useful thus improving utilisation of overall model capacity and the second aims at distributing nodes uniformly across assigned clusters. To achieve these goals a second regularisation term is defined as follows:
\begin{align}
\label{eq:L_C}
\begin{split}
L_{C} &= \sum_{l} L_C^{(l)} \\
L_{C}^{(l)} &= H(\frac{1}{n_l} \mathbf{1}^T S^{(l)})
\end{split}
\end{align}

Where $\mathbf{1}$ is a column vector of $1$ with dimension $n_l$. Combining $L_E$ and $L_C$ reduces the solution space drastically and obtains a minimum where nodes are assigned "uniquely" to clusters yet are spread uniformly across all available clusters. \newline

\textbf{Removal of link prediction objective}. The authors of \cite{NIPS2018_7729} explain that the auxiliary link prediction objective and and its corresponding loss term $L_{LP} = \norm{A^{(l)},S^{(l)}S^{(l)^T}}_F$ where $\norm{\cdot}_F$ denotes the Frobenius norm, was introduced to encode the intuition that nearby nodes should be pooled together. Furthermore it seems plausible that this auxiliary task also prevent $L_E$ from collapsing the training process in an early stage into a spurious local minima where at most one or two clusters are utilised and hence limiting the capacity of the model. Since the structural similarity features encode the the notion that neighbouring nodes have features that are "close" and the introduced regularisation term $L_C$ encourages utilising the full model capacity it seems that the link prediction objective is now redundant.

\section{Experimental Results}
\label{S:Experimental Results}
\subsection{Goals}
\label{S:Experimental Results:Goals}
A number of inductive graph classification tasks are performed in order to evaluate the effectiveness of the structural similarity features and effect of proposed changes to DiffPool with the goal of answering the following questions:
\begin{enumerate}
\item How does the use of structural similarity features compare to the use of node features for calculating cluster assignments?
\item What is the effect of information and permutation invariance loss as the result of utilising only the ordered indices mapping of top-$k$ structural similarity features due to the application of the indices mapping trick?
\item What is the effect of increasing the parameter $p$, the power of the adjacency matrix?
\item What is the effect of the proposed regularisation term $L_C$?
\item How does SimPool compare to other recently proposed methods for hierarchical graph pooling on graph classification tasks?
\end{enumerate}

\subsection{Datasets}
\label{S:Experimental Results:Datasets}
The benchmark datasets used are summarised in table \ref{tab:datasets}. These datasets are revised versions of common benchmarks used in graph classification tasks where isomorphic graphs have been removed \cite{ivanov2019understanding} \footnote{Datasets are available at \url{https://github.com/nd7141/graph_datasets}}.

\begin{table}[t]
\centering
\begin{tabular}{l c c c c} 
\toprule
Dataset & Size & Avg. Nodes & Avg. Edges & Classes \\ [0.5ex] 
\midrule
Enzymes \cite{Borgwardt_2005} & 595 & 32.48 & 62.17 & 6 \\ 
D\&D \cite{Dobson_2003} & 1178 & 284.32 & 715.66 & 2 \\ 
\bottomrule
\end{tabular}
\caption{\label{tab:datasets}Summary of datasets used for inductive learning graph classification experiments.} 
\end{table}

\subsection{Model architecture}
\label{S:Experimental Results:Model architecture}
All experiments share a common simple feed forward model architecture however layer parameters and inputs are modified per experiment as appropriate. The architecture and parameters used for each of the datasets are summarised in tables \ref{tab:architecture} and \ref{tab:parameters}, these values are generally unchanged unless explicitly noted. Furthermore no experimentation with more complex architectures, intermediate readout layers or skip connections was performed. For optimisation Adam \cite{DBLP:journals/corr/KingmaB14} with the default parameter values is employed and a learning rate of $10^{-4}$. Note that there was no attempt to find an optimal architecture or hyperparameter settings for the experiments but merely to measure the relative accuracy for different hyperparameter settings and between the different feature types under this basic network configuration. The indices mapping trick was only applied in the calculation of $S^{(0)}$, for calculating $S^{(1)}$ the structural similarity features were used as is. Lastly, $S^{(2)}= \mathbf{1}$, and $X^{(l)}$, $A^{(l)}$ are calculated as per equation \eqref{eq:diffpool}.

\begin{table}[t]
\centering
\begin{tabular}{l l l c c l}
\toprule
\multirow{2}{*}{Output} & \multirow{2}{*}{Module} & \multirow{2}{*}{Layers} & \multicolumn{2}{c}{Units} & \multirow{2}{*}{Activation} \\ [0.5ex] \cline{4-5}
 &  &  & \multicolumn{1}{c}{Enzymes} & \multicolumn{1}{c}{D\&D} & \\ [0.5ex]
\midrule
\multirow{3}{*}{$Z^{(0)}$} & \multicolumn{1}{l}{GMN encoder} & \multicolumn{1}{l}{dense} & \multicolumn{1}{c}{512} & \multicolumn{1}{c}{1024} & \multicolumn{1}{l}{ReLU \citep{10.5555/3104322.3104425}} \\ 
& \multicolumn{1}{l}{GMN propagation} & \multicolumn{1}{l}{dense} & \multicolumn{1}{c}{512} & \multicolumn{1}{c}{1024} & \multicolumn{1}{l}{ReLU} \\ 
& \multicolumn{1}{l}{GMN propagation} & \multicolumn{1}{l}{dense} & \multicolumn{1}{c}{512/256} & \multicolumn{1}{c}{1024} & \multicolumn{1}{l}{linear} \\ 

\hline

\multirow{3}{*}{$S^{(0)}$} & \multicolumn{1}{l}{GMN encoder} & \multicolumn{1}{l}{dense} & \multicolumn{1}{c}{512} & \multicolumn{1}{c}{1024} & \multicolumn{1}{l}{ReLU} \\ 
& \multicolumn{1}{l}{GMN propagation} & \multicolumn{1}{l}{dense} & \multicolumn{1}{c}{512} & \multicolumn{1}{c}{1024} & \multicolumn{1}{l}{ReLU} \\ 
& \multicolumn{1}{l}{GMN propagation} & \multicolumn{1}{l}{dense} & \multicolumn{1}{c}{512/8} & \multicolumn{1}{c}{1024/32} & \multicolumn{1}{l}{linear/softmax} \\ 

\hline

\multirow{1}{*}{$Z^{(1)}$} & \multirow{1}{*}{GCN} & & \multicolumn{1}{c}{512} & \multicolumn{1}{c}{2048} & \multicolumn{1}{l}{ReLU} \\ 

\hline

\multirow{2}{*}{$S^{(1)}$} & \multirow{2}{*}{MLP} & \multirow{1}{*}{dense} & \multicolumn{1}{c}{256} & \multicolumn{1}{c}{2048} & \multicolumn{1}{l}{ReLU} \\
 & & \multirow{1}{*}{dense} & \multicolumn{1}{c}{4} & \multicolumn{1}{c}{8} & \multicolumn{1}{l}{softmax} \\ 

\hline

\multirow{1}{*}{$Z^{(2)}$} & \multirow{1}{*}{GCN} & & \multicolumn{1}{c}{1024} & \multicolumn{1}{c}{4096} & \multicolumn{1}{l}{ReLU} \\ 

\hline

\multirow{1}{*}{predictions} & \multirow{1}{*}{MLP} & \multirow{1}{*}{dense} & \multicolumn{1}{c}{6} & \multicolumn{1}{c}{2} & \multicolumn{1}{l}{softmax}\\ 

\bottomrule
\end{tabular}
\caption{\label{tab:architecture}Common architecture and parameters used for all experiments. For the GMN modules when a single value is specified for the units it refers to both $f_{message}$ and $f_{node}$} 
\end{table}

\begin{table}[t]
\centering
\begin{tabular}{l c c c c c c c c} 
\toprule
Dataset & $p$ & $k^{(0)}$ & $\alpha$ & $\lambda$ & $L_E$ scaler & $L_C$ scaler & tra/val split & epochs\\ [0.5ex] 
\midrule
Enzymes &  \multirow{2}{*}{1} & 12 &  \multirow{2}{*}{1.0} & \multirow{2}{*}{0} & 1.0 & \multirow{2}{*}{1.0} & \multirow{2}{*}{0.9/0.1} & 100\\ 
D\&D & & 25 & & & 0.4 & & & 230\\ 
\bottomrule
\end{tabular}
\caption{\label{tab:parameters}Default parameter values used for all experiments by dataset where not stated explicitly otherwise. These values are likely not optimal but rather chosen as experimental baselines. The Epochs columns indicates the number of epochs used for training.}
\end{table}

\subsection{Assignment features}
\label{S:Experimental Results:Assignment features}
For evaluating the utility of the structural similarity features a graph classification task is repeated with the same model architecture and parameters where the only variant is the features used for calculating the assignment matrix $S$. There are three variants compared: structural similarity features, node features and concatenation of both. 

\begin{figure}[h!]
\centering
\begin{subfigure}{.5\textwidth}
    \centering
    \includegraphics[width=1.0\textwidth]{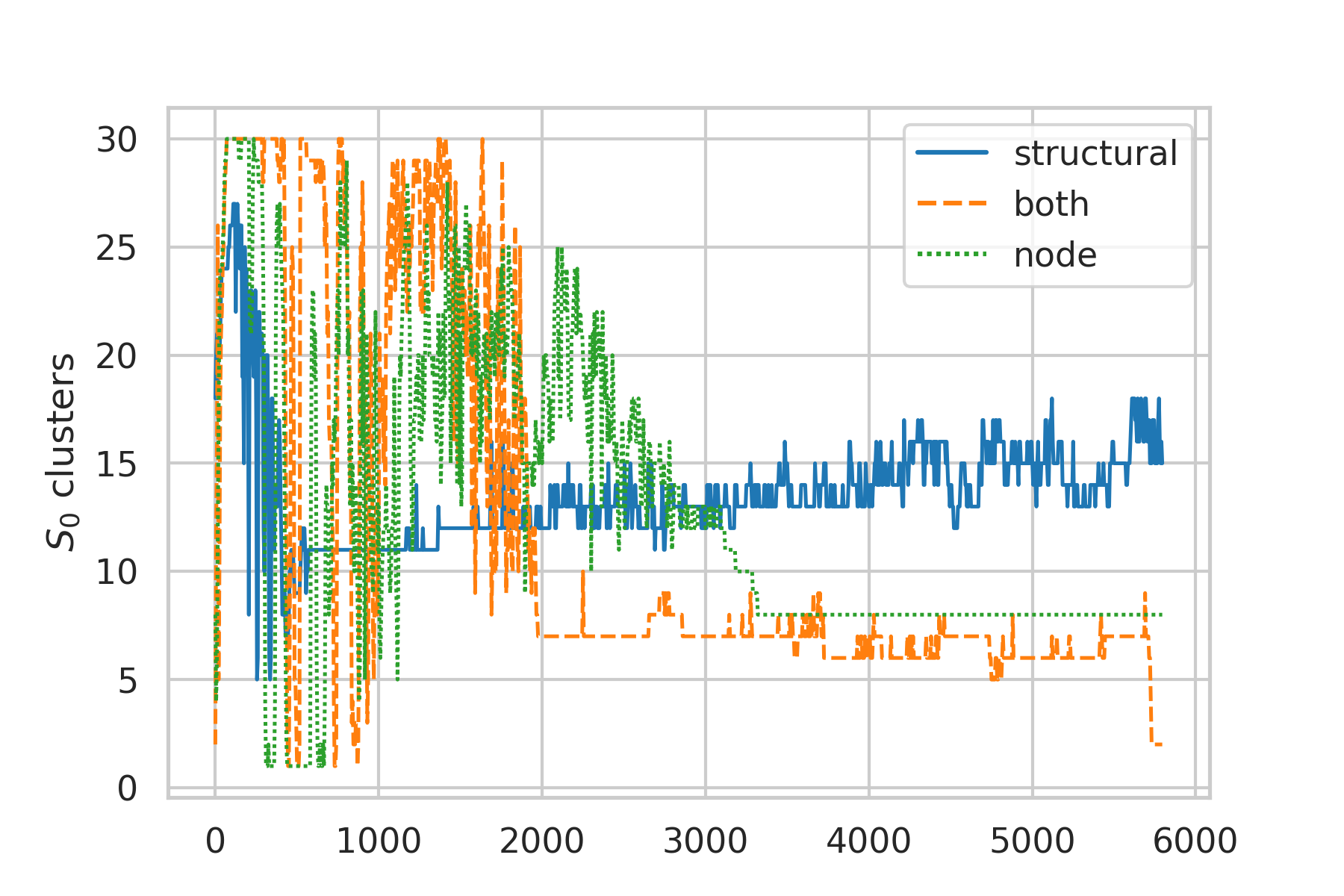}
    \caption{}
\end{subfigure}%
\begin{subfigure}{.5\textwidth}
    \centering
    \includegraphics[width=1.0\textwidth]{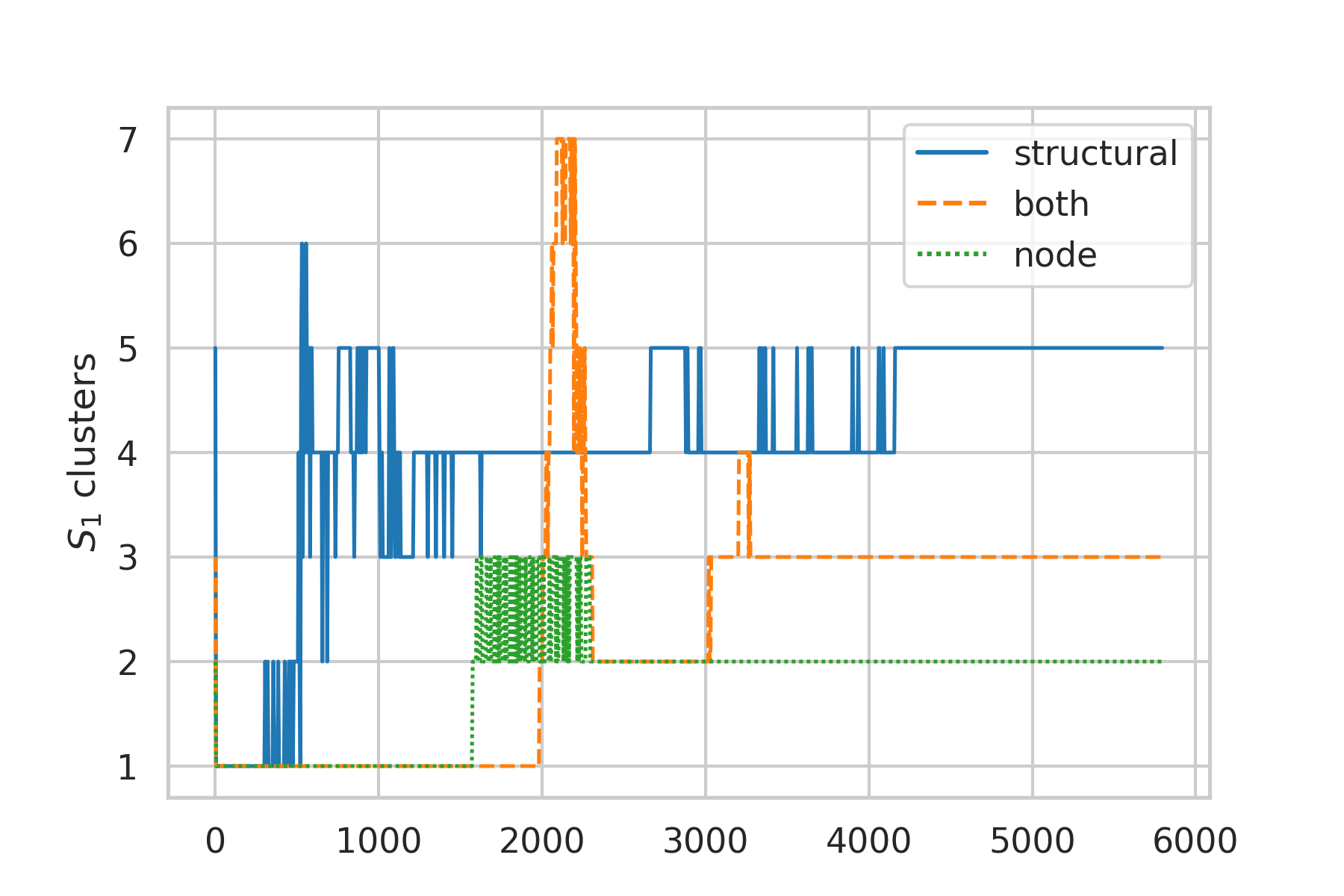}
    \caption{}
\end{subfigure}
\begin{subfigure}{.5\textwidth}
    \centering
    \includegraphics[width=1.0\textwidth]{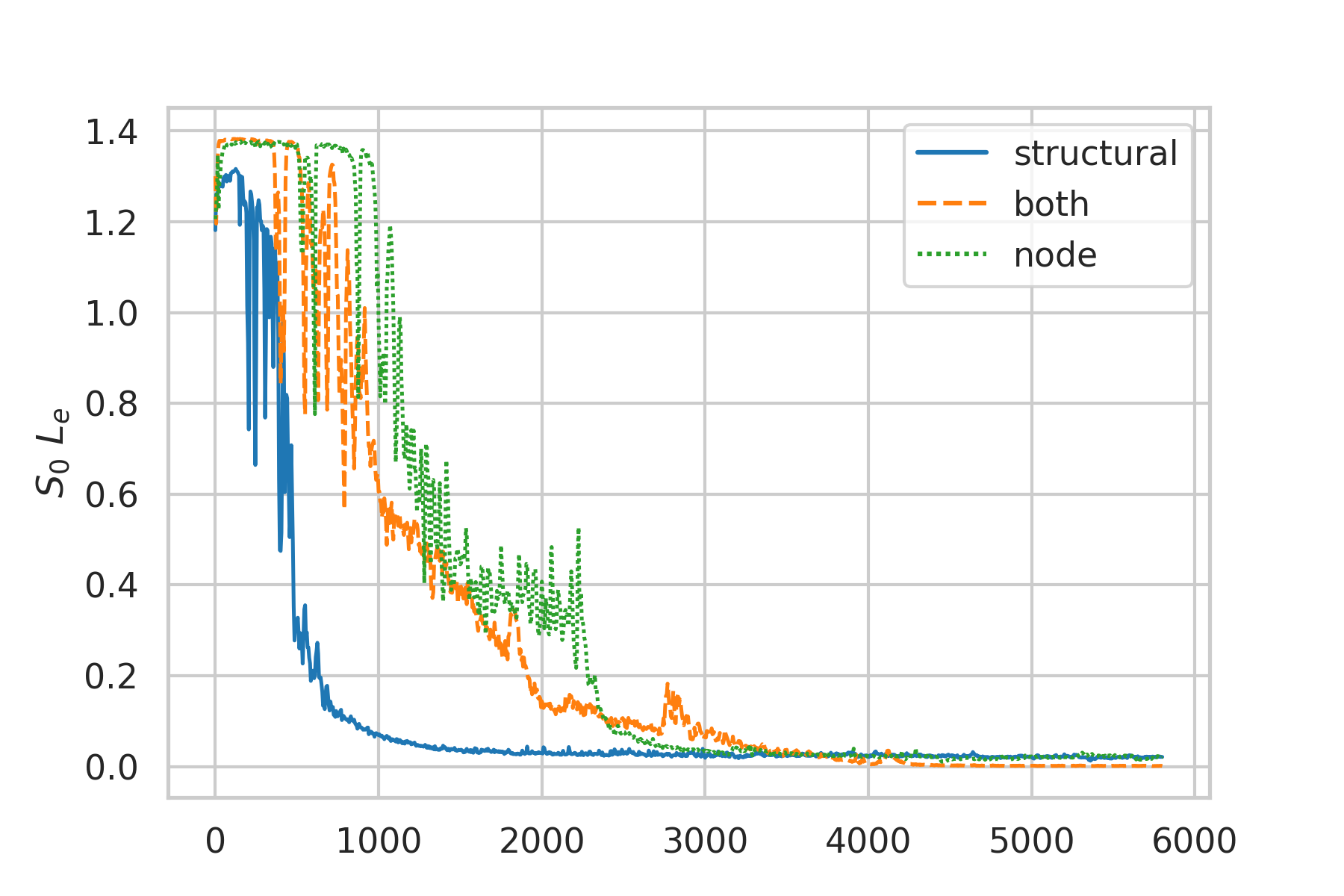}
    \caption{}
\end{subfigure}%
\begin{subfigure}{.5\textwidth}
    \centering
    \includegraphics[width=1.0\textwidth]{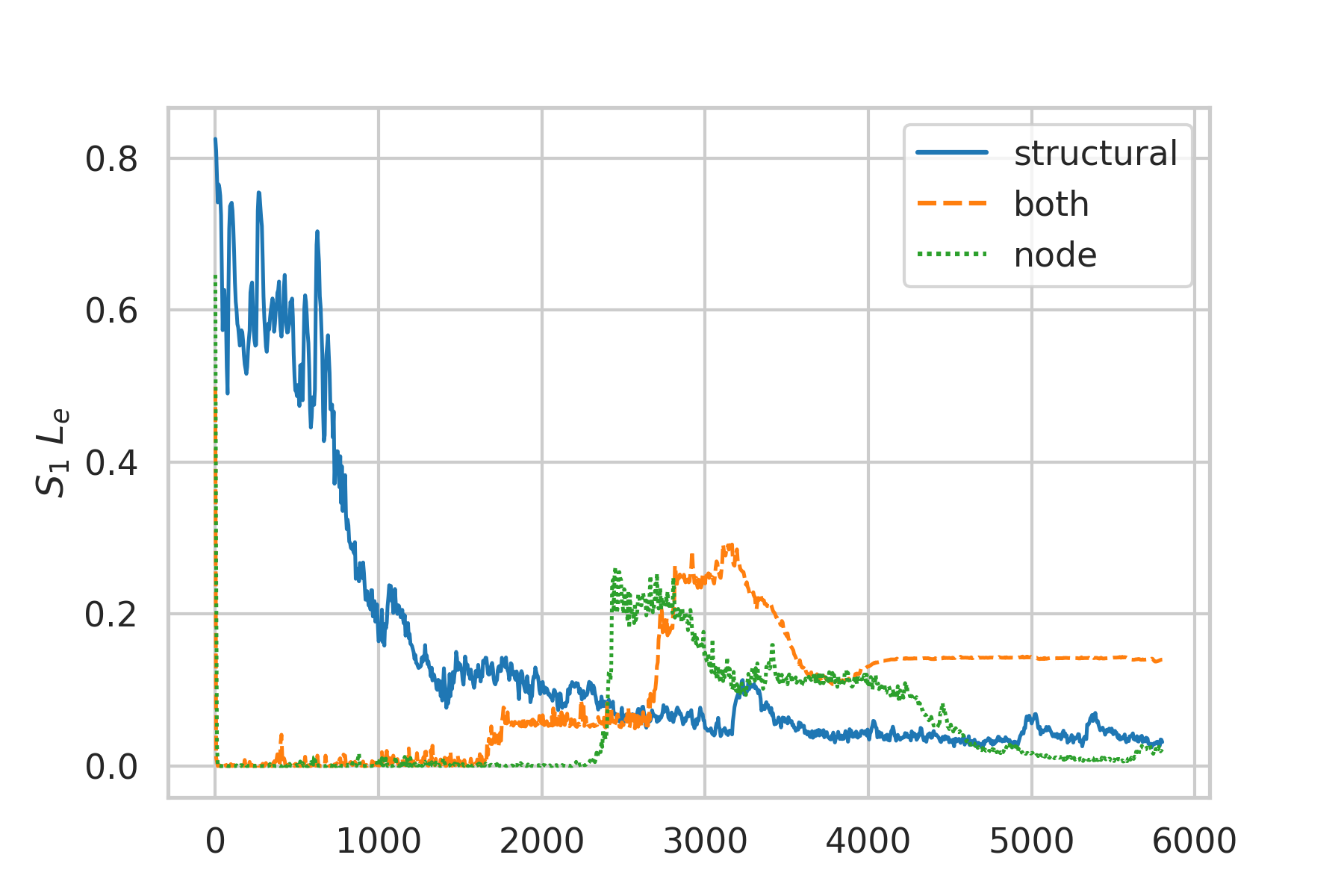}
    \caption{}
\end{subfigure}
\begin{subfigure}{.5\textwidth}
    \centering
    \includegraphics[width=1.0\textwidth]{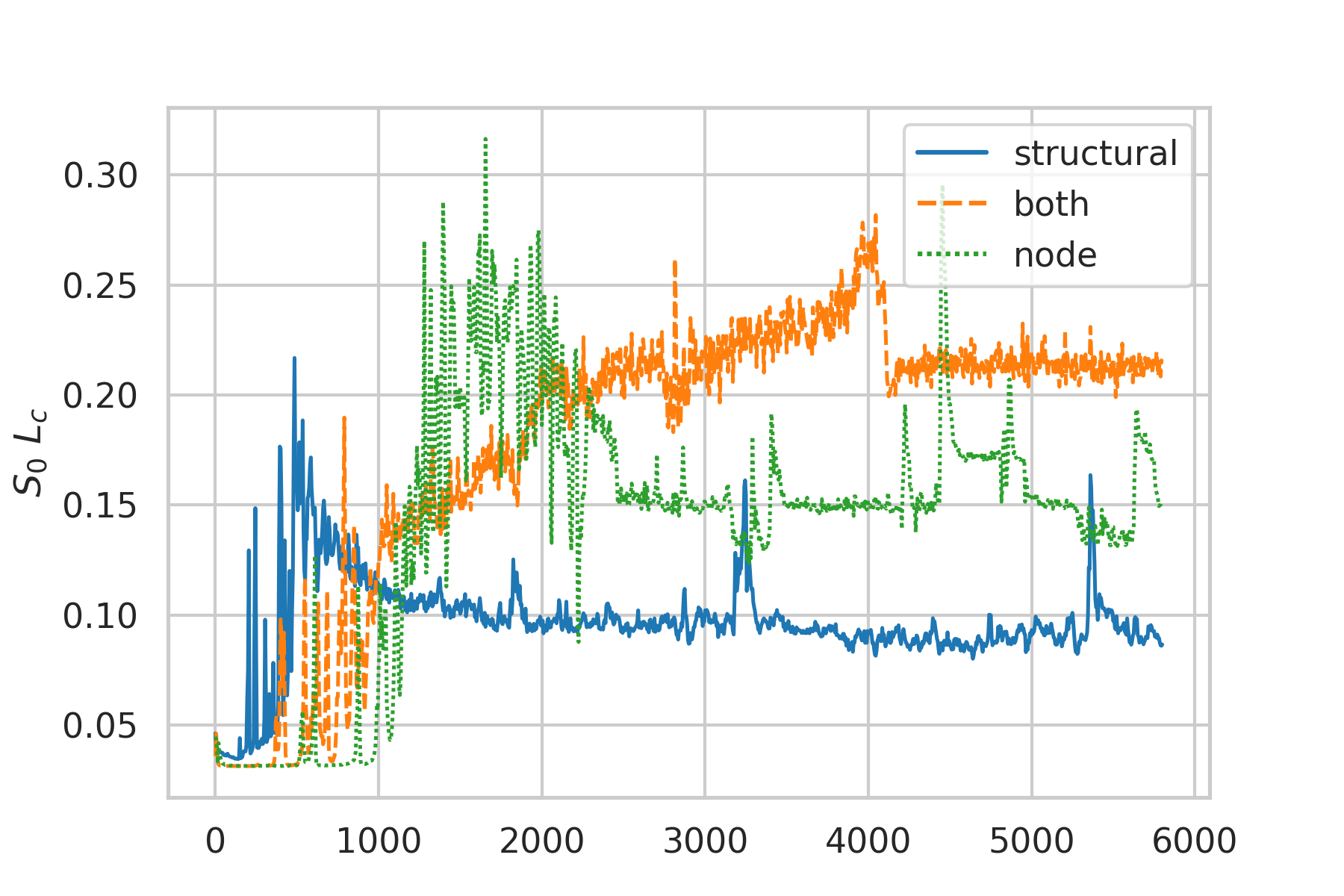}
    \caption{}
\end{subfigure}%
\begin{subfigure}{.5\textwidth}
    \centering
    \includegraphics[width=1.0\textwidth]{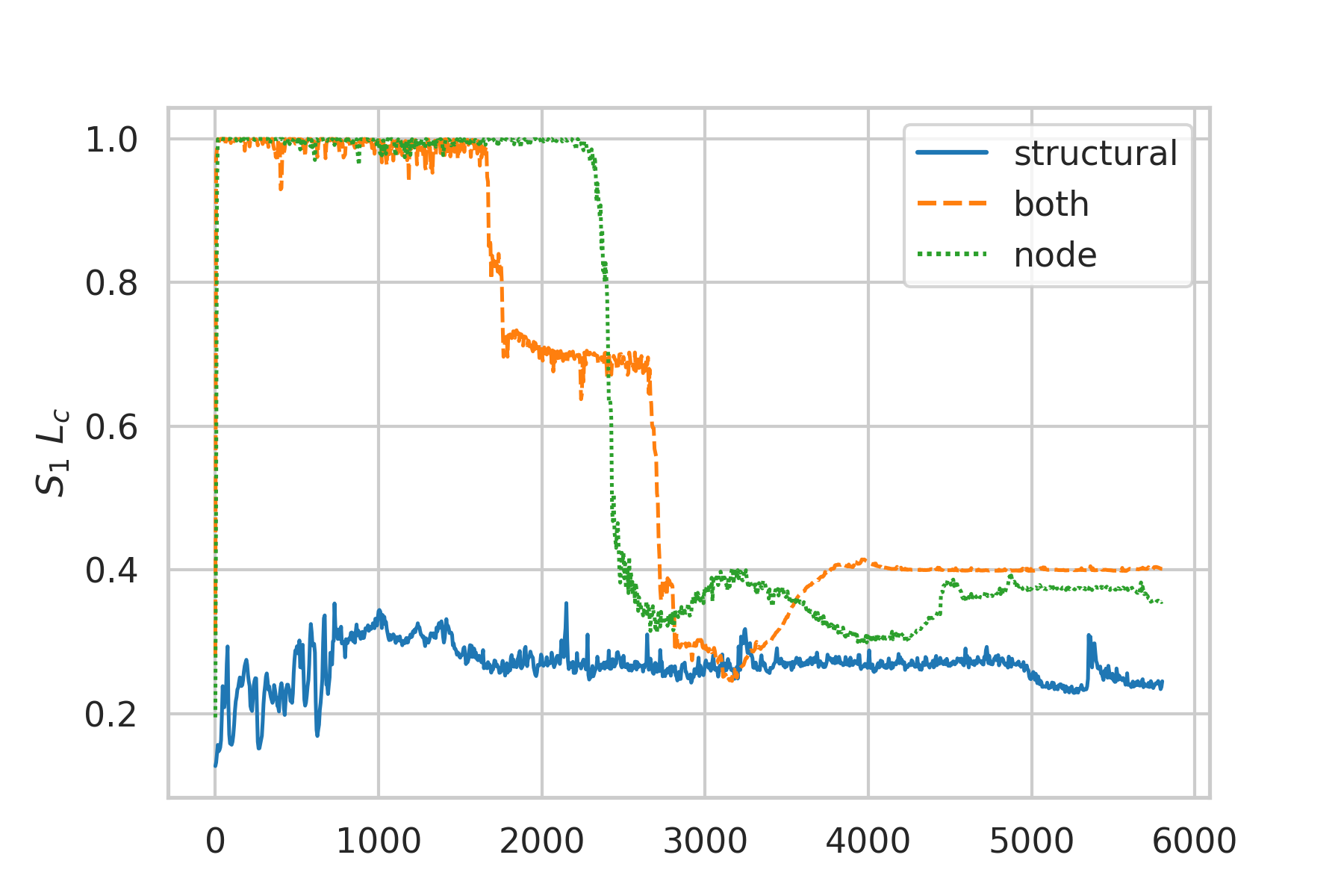}
    \caption{}
\end{subfigure}
\caption[short]{Training statistics of the D\&D dataset for both assignment matrices. $S^{(0)}$ (left column) and $S^{(1)}$ (right column). (a) and (b) number of different clusters selected; (c) and (d) loss term $L_E$ representing the uniqueness of nodes cluster assignments; (e) and (f) loss term $L_C$ representing the uniformity in cluster assignment as well as utilisation of all available clusters.}
\label{fig:DD num clusters}
\end{figure}

\begin{figure}[h!]
\centering
\begin{subfigure}{.5\textwidth}
    \centering
    \includegraphics[width=1.0\textwidth]{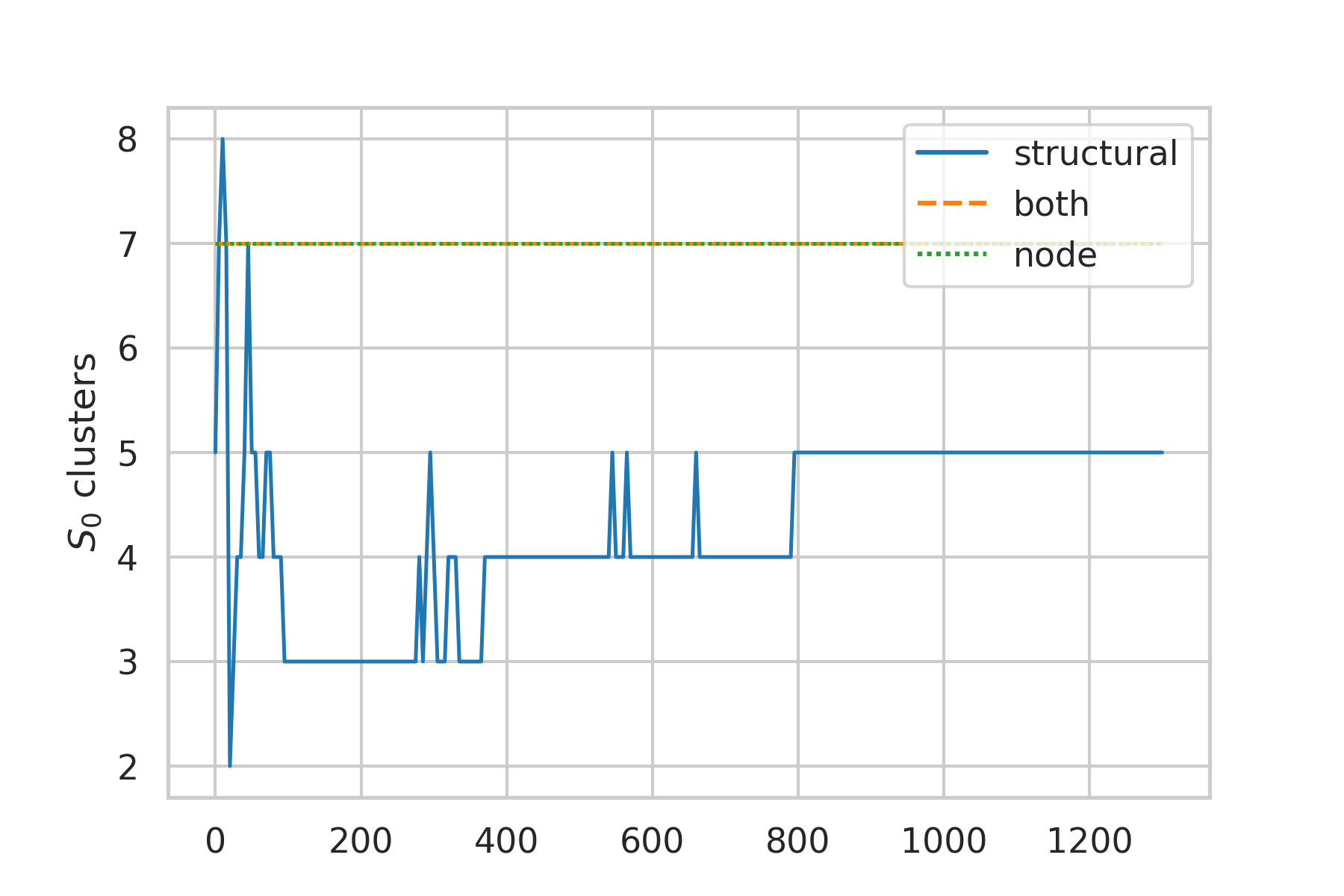}
    \caption{}
\end{subfigure}%
\begin{subfigure}{.5\textwidth}
    \centering
    \includegraphics[width=1.0\textwidth]{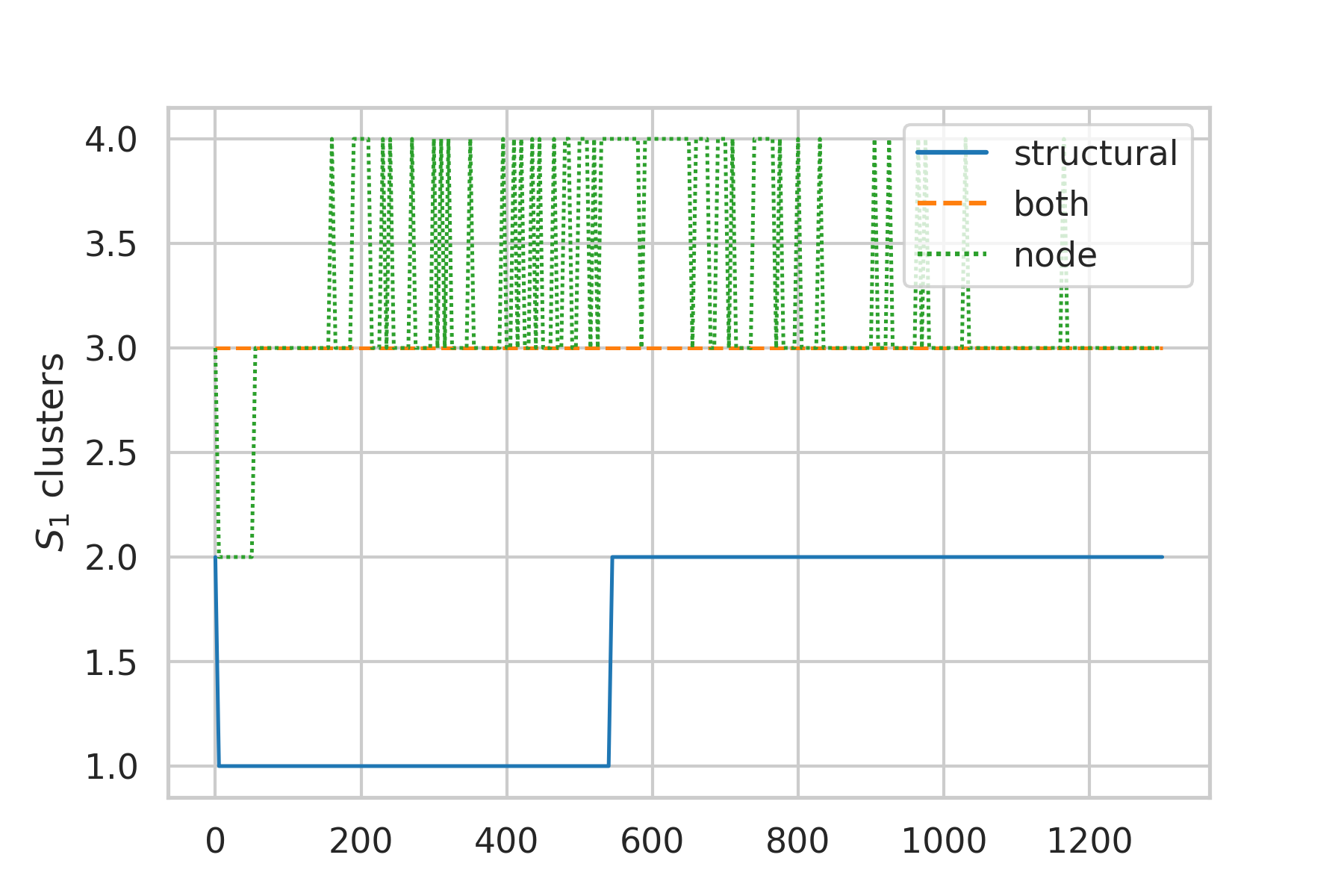}
    \caption{}
\end{subfigure}
\begin{subfigure}{.5\textwidth}
    \centering
    \includegraphics[width=1.0\textwidth]{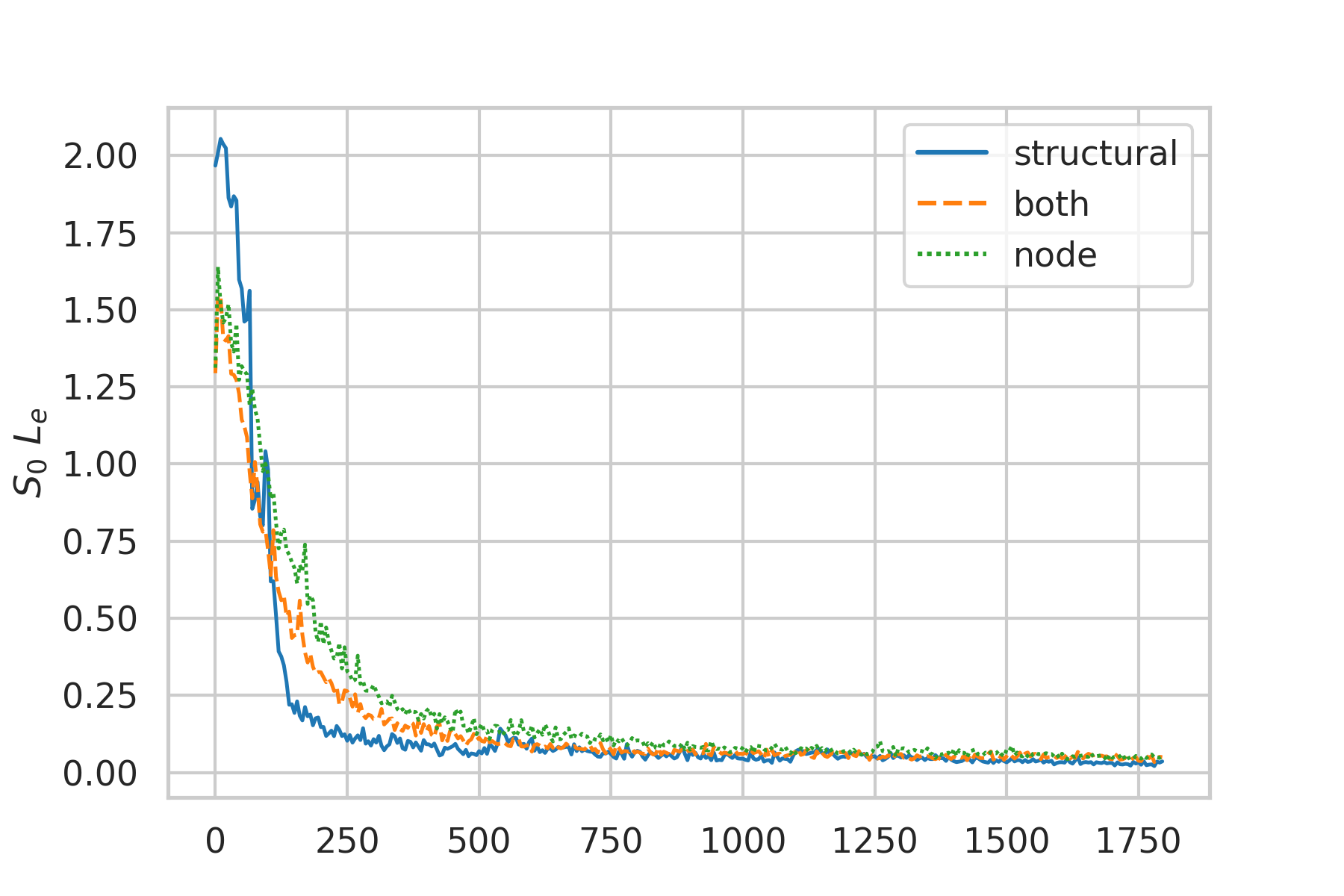}
    \caption{}
\end{subfigure}%
\begin{subfigure}{.5\textwidth}
    \centering
    \includegraphics[width=1.0\textwidth]{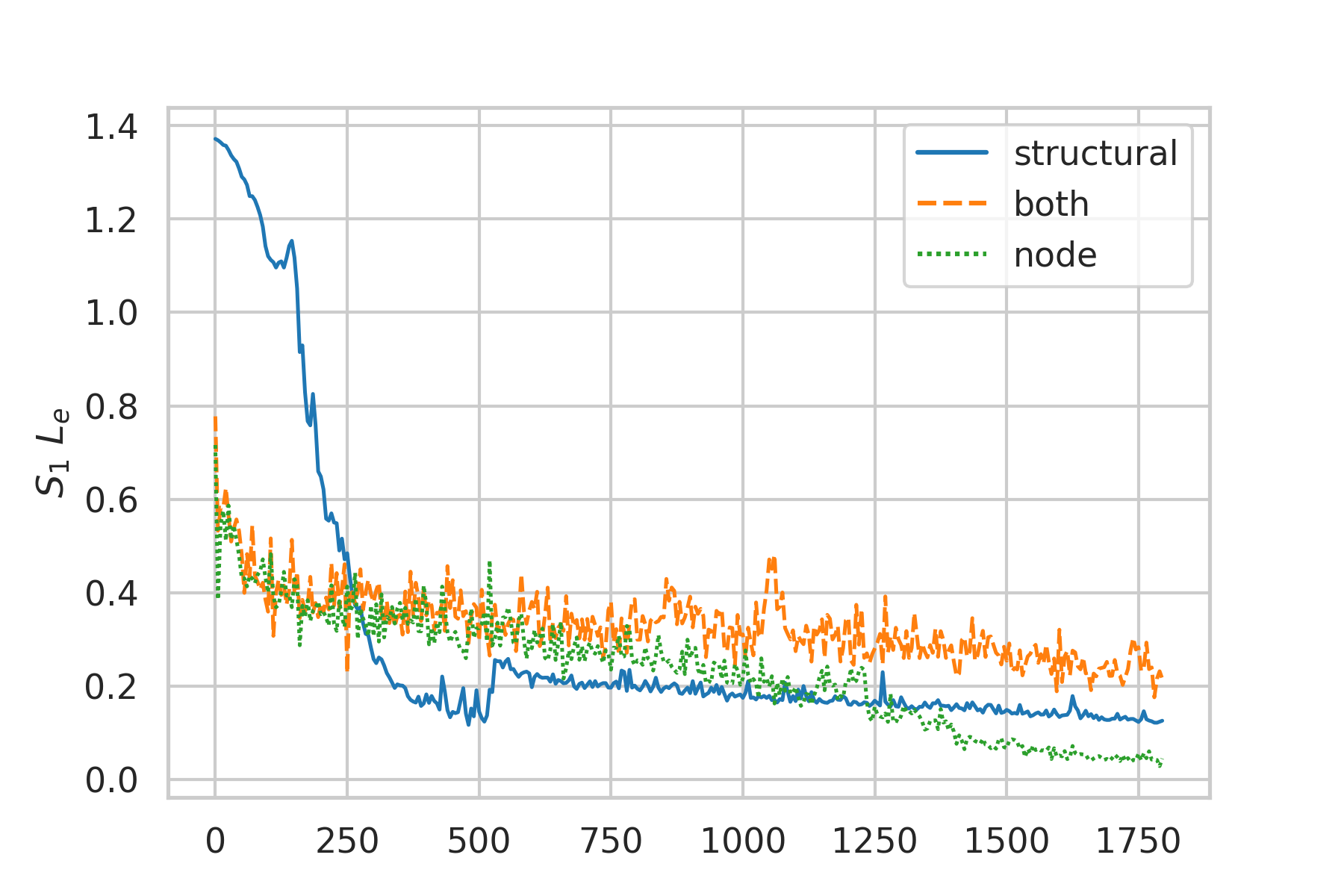}
    \caption{}
\end{subfigure}
\begin{subfigure}{.5\textwidth}
    \centering
    \includegraphics[width=1.0\textwidth]{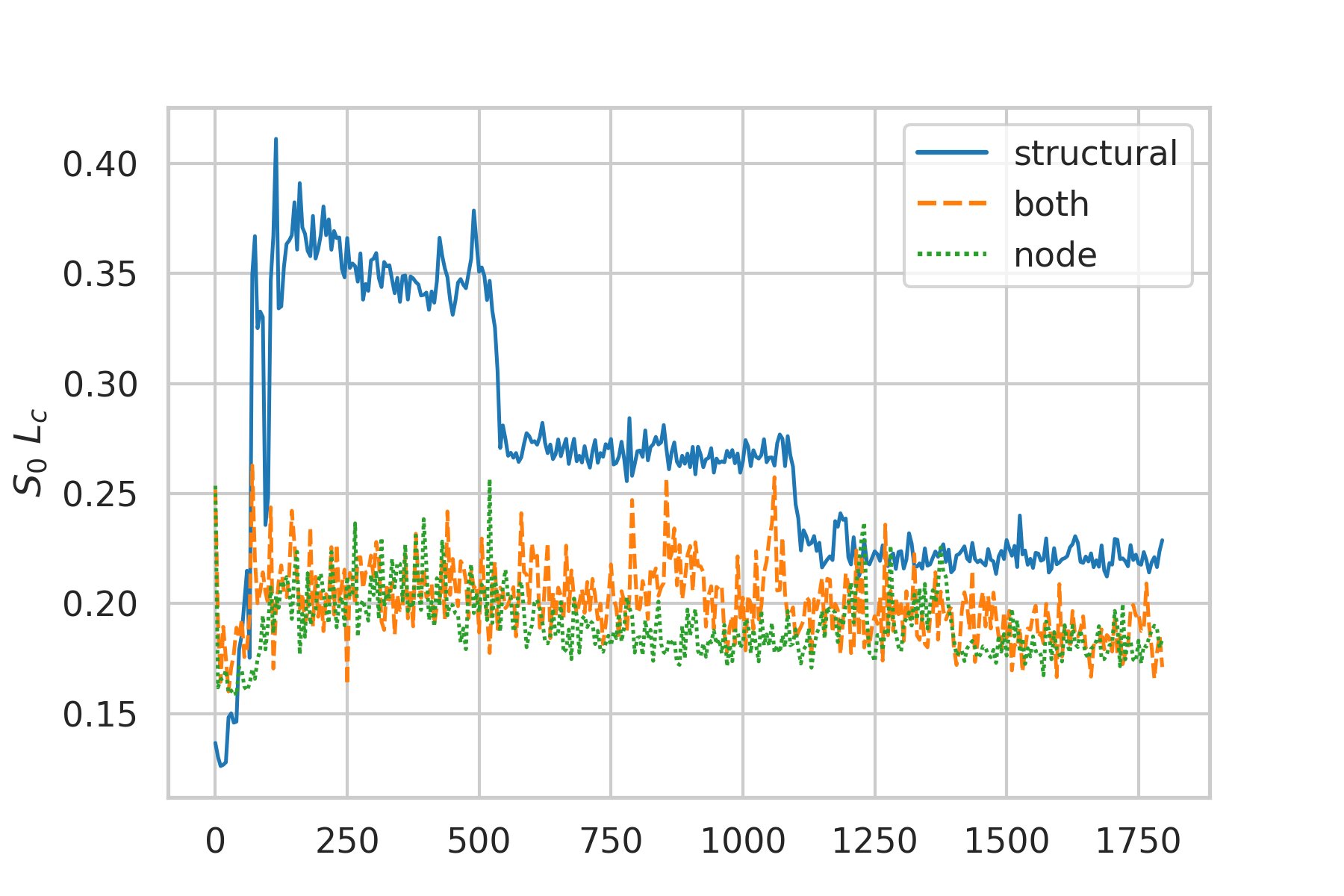}
    \caption{}
\end{subfigure}%
\begin{subfigure}{.5\textwidth}
    \centering
    \includegraphics[width=1.0\textwidth]{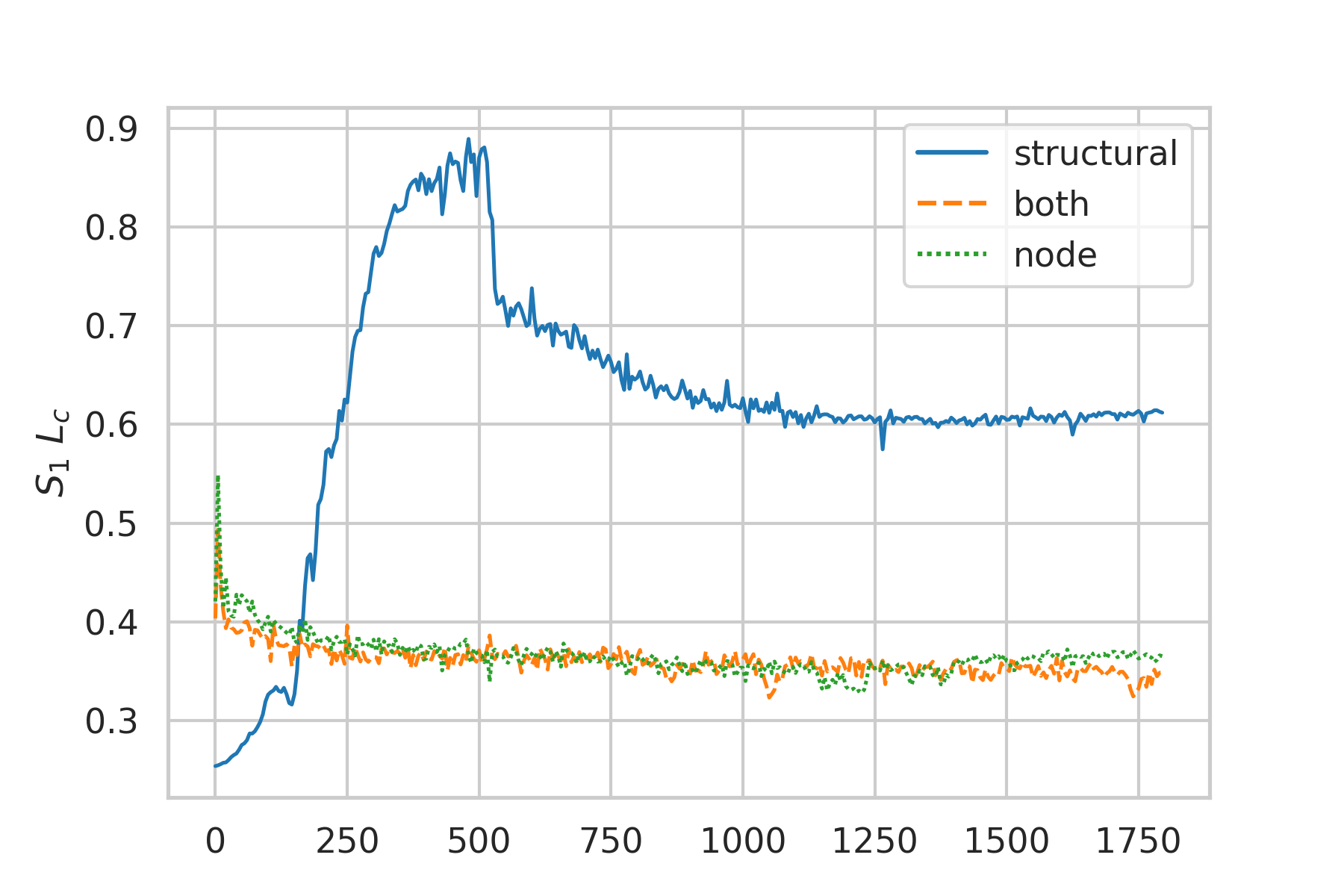}
    \caption{}
\end{subfigure}
\caption[short]{Training statistics of the Enzymes dataset for both assignment matrices. $S^{(0)}$ (left column) and $S^{(1)}$ (right column). (a) and (b) number of different clusters selected; (c) and (d) loss term $L_E$ representing the uniqueness of nodes cluster assignments; (e) and (f) loss term $L_C$ representing the uniformity in cluster assignment as well as utilisation of all available clusters.}
\label{fig:enzymes num clusters}
\end{figure}

Figures \ref{fig:DD num clusters} and \ref{fig:enzymes num clusters} illustrate values of the loss terms $L_E$ and $L_C$ during training and the number of different clusters utilised where a node cluster assignment is determined by the $argmax$ of the corresponding row in the assignment matrix. Generally a combination of low values for both loss terms represents a desirable policy of cluster assignment that is both uniform across all available clusters and unique for each node. The figures indicate that in these experiments the "node" and "both" variants seem to result in similar training pattens whereas using only the structural features for calculating assignments allows the model to choose a different and distinct training path. In the case of the D\&D dataset using only the structural similarity features for cluster assignments enables the model to utilise substantially more clusters at both pooling layers. It also appears that training is more stable and assignments are becoming distinct in earlier stages of training despite utilising more clusters effectively. Furthermore using the "structural" variant consistently attains the maximal accuracy obtained on the validation data as summarised in table \ref{tab:max accuracy}. 

\begin{figure}[h!]
\centering
\begin{subfigure}{.5\textwidth}
    \centering
    \includegraphics[width=1.0\textwidth]{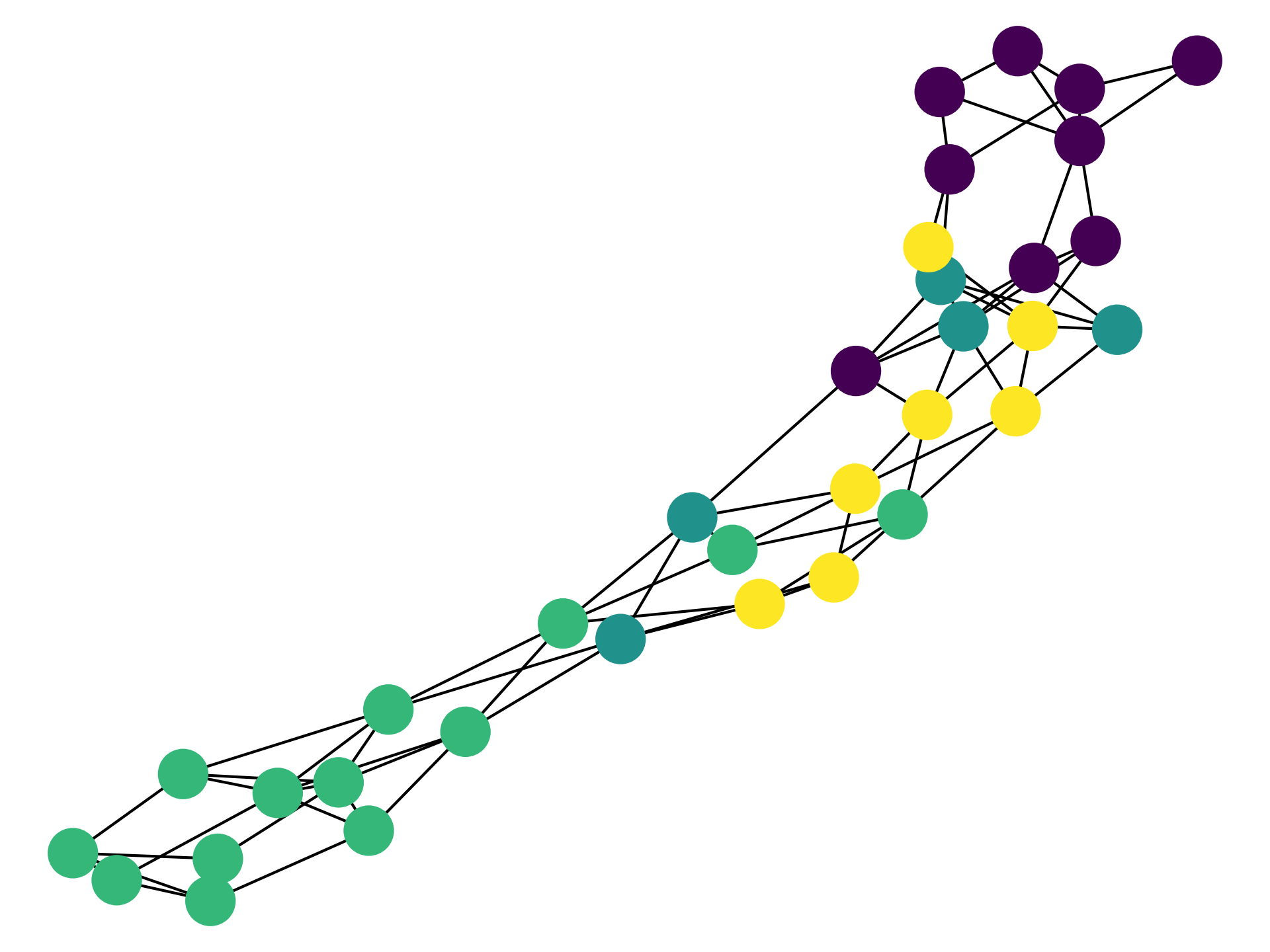}
    \caption{}
\end{subfigure}%
\begin{subfigure}{.5\textwidth}
    \centering
    \includegraphics[width=1.0\textwidth]{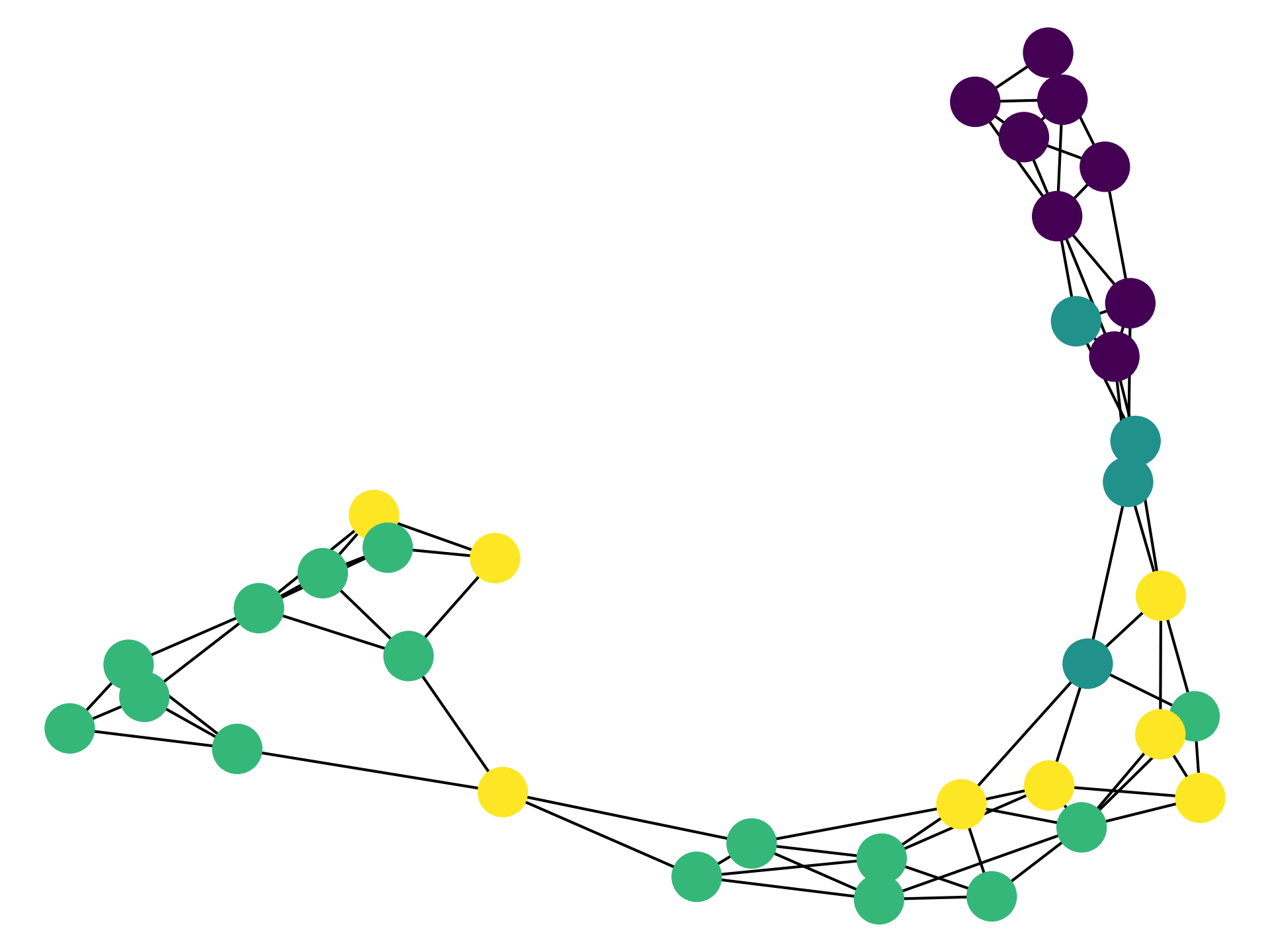}
    \caption{}
\end{subfigure}
\begin{subfigure}{.5\textwidth}
    \centering
    \includegraphics[width=1.0\textwidth]{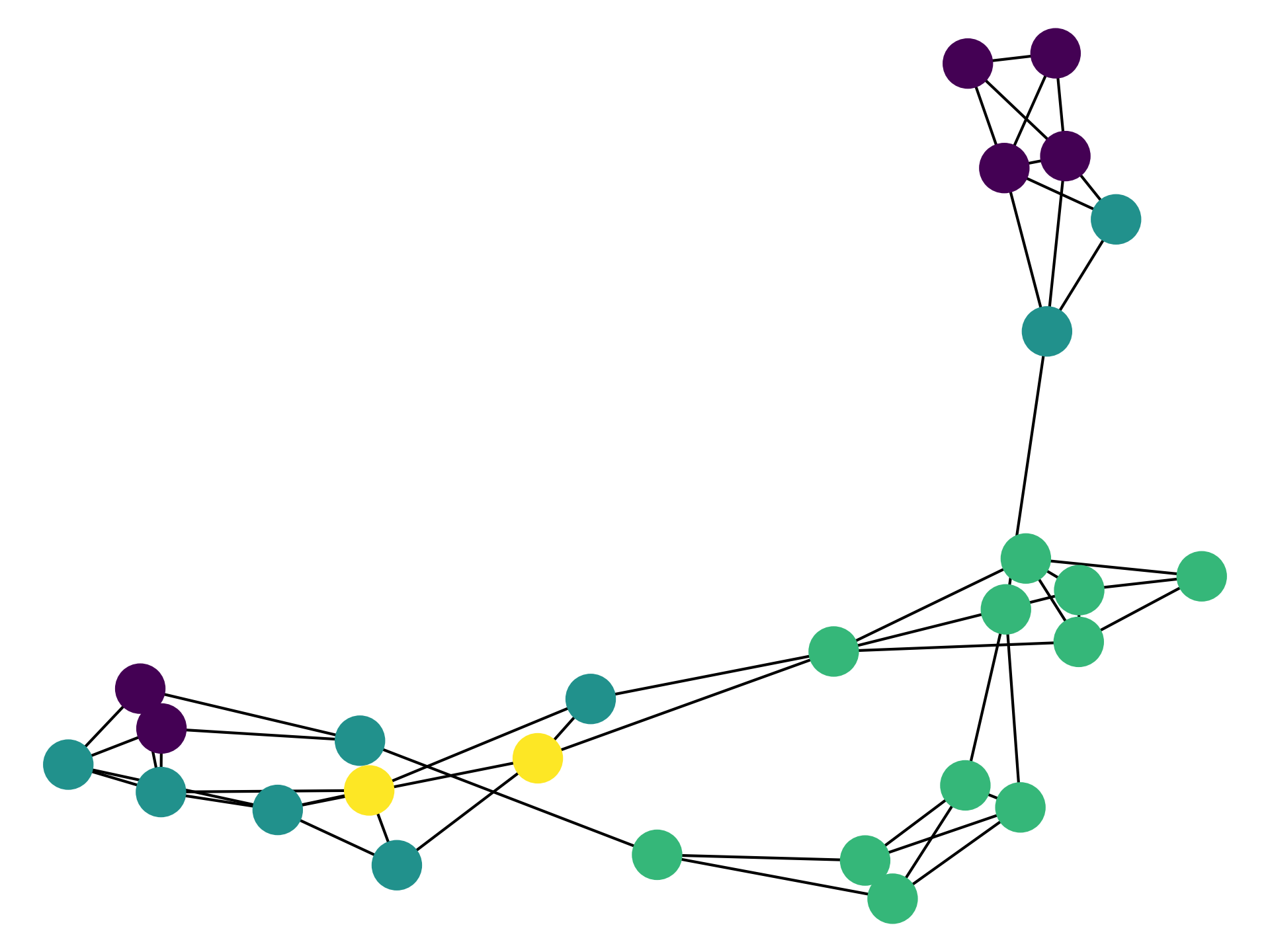}
    \caption{}
\end{subfigure}%
\begin{subfigure}{.5\textwidth}
    \centering
    \includegraphics[width=1.0\textwidth]{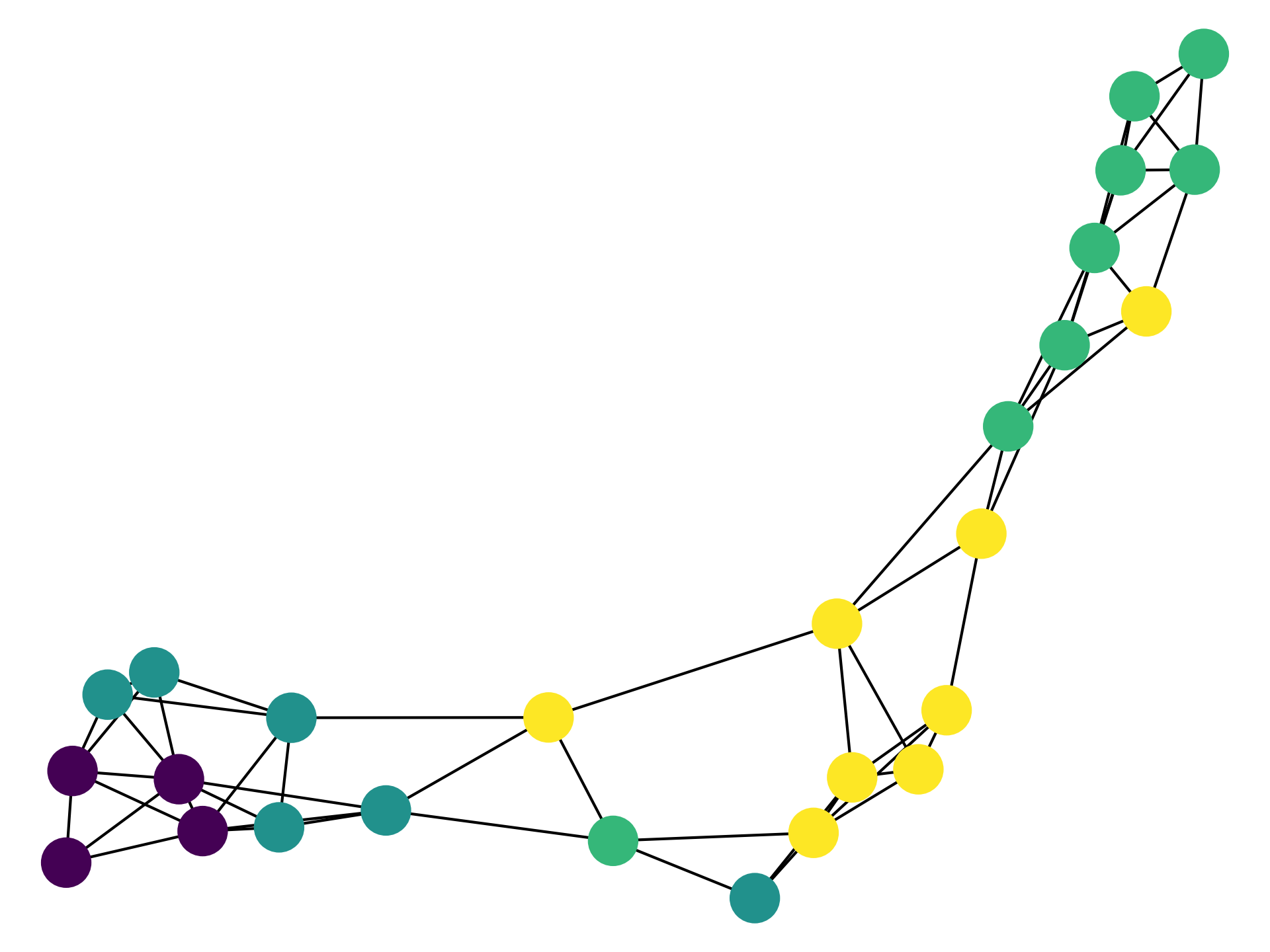}
    \caption{}
\end{subfigure}
\begin{subfigure}{.5\textwidth}
    \centering
    \includegraphics[width=1.0\textwidth]{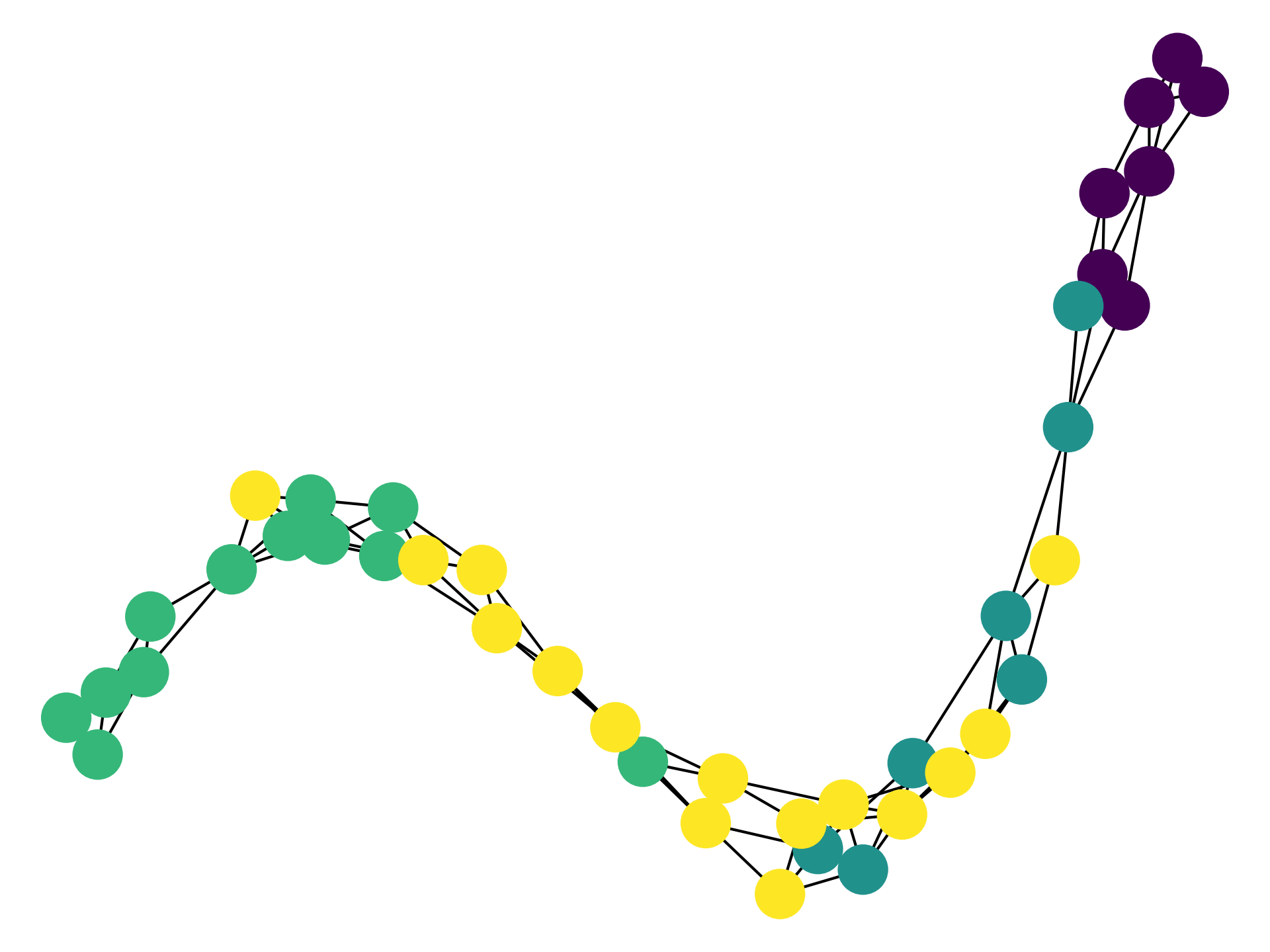}
    \caption{}
\end{subfigure}%
\begin{subfigure}{.5\textwidth}
    \centering
    \includegraphics[width=1.0\textwidth]{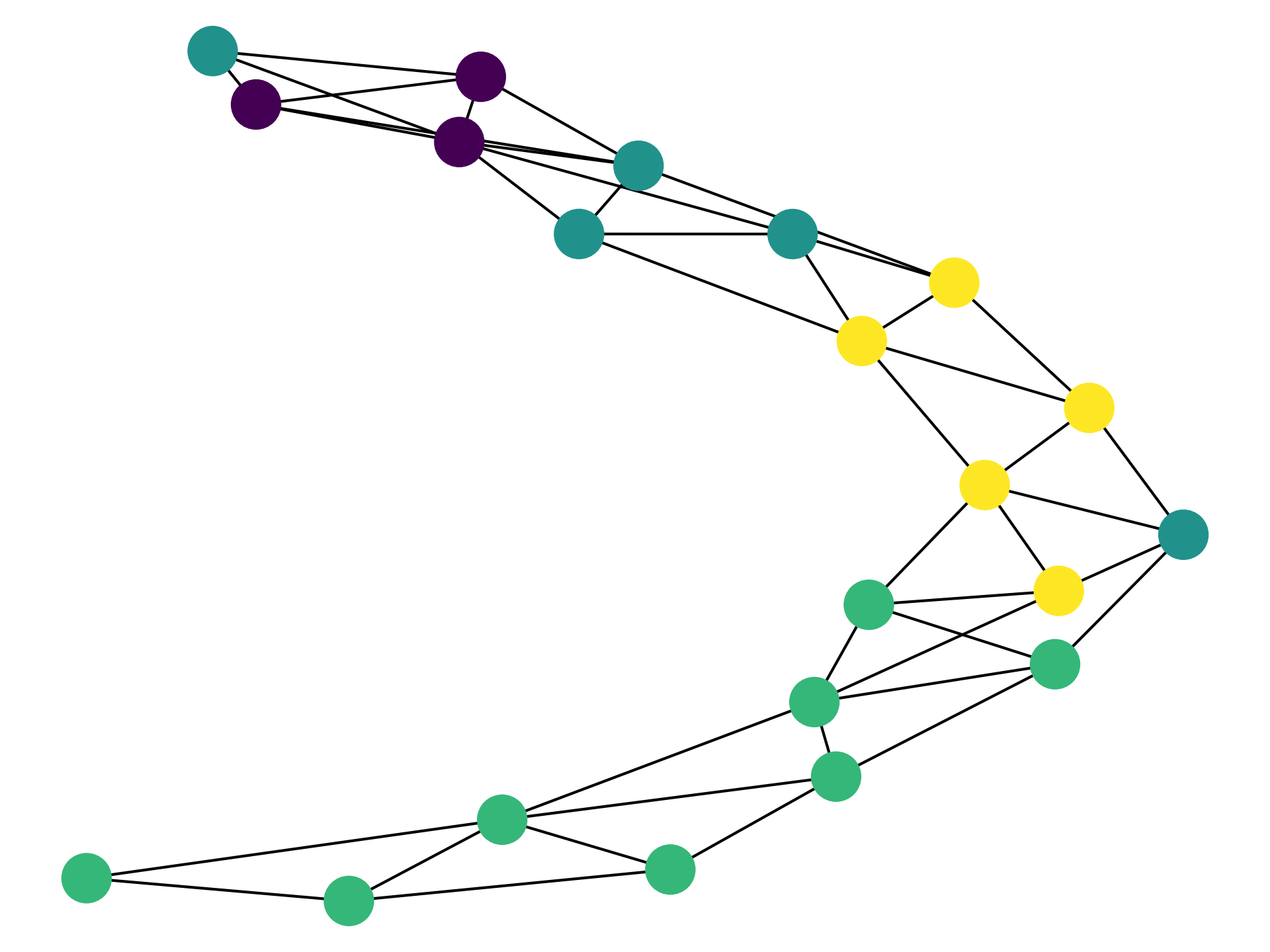}
    \caption{}
\end{subfigure}
\begin{subfigure}{.5\textwidth}
    \centering
    \includegraphics[width=1.0\textwidth]{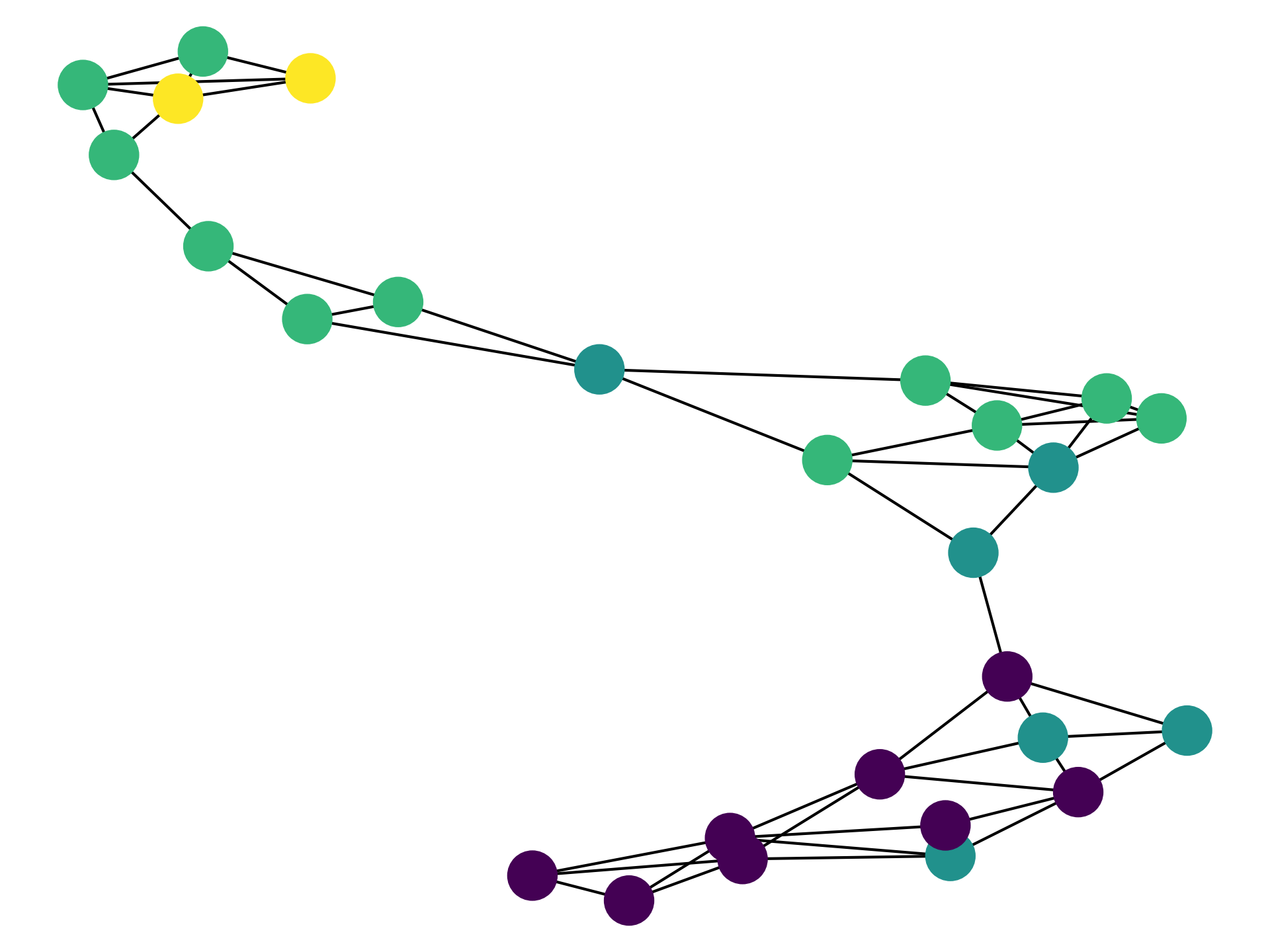}
    \caption{}
\end{subfigure}%
\begin{subfigure}{.5\textwidth}
    \centering
    \includegraphics[width=1.0\textwidth]{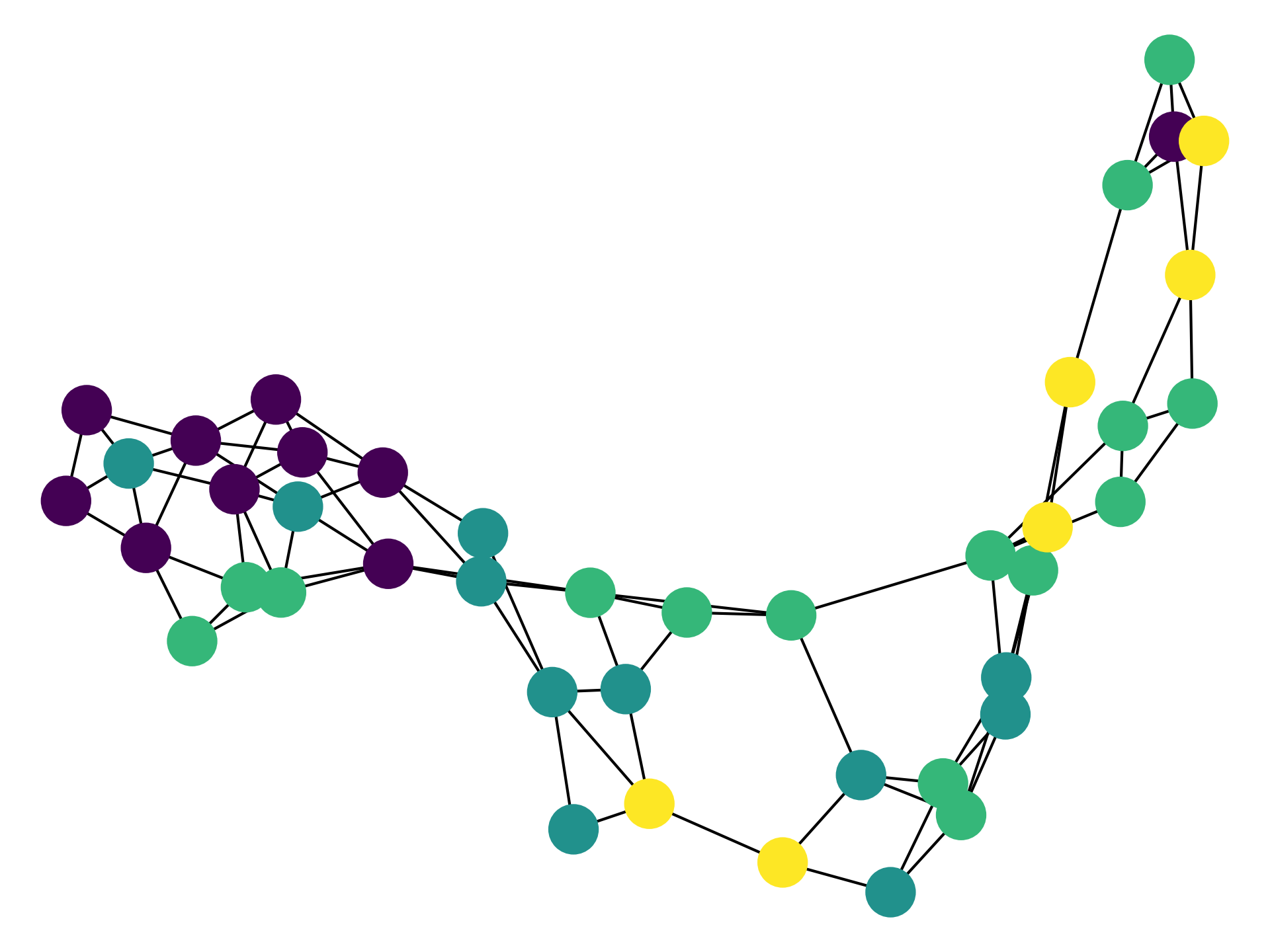}
    \caption{}
\end{subfigure}
\caption[short]{Examples of node clustering calculated by the first pooling layer using the structural similarity features for graphs in the Enzymes dataset with $\lambda=0, \, \alpha=0.8$ and $k=12$. Node colors represent cluster assignments determined by the $argmax$ of the node assignment distribution.}
\label{fig:ssf cluster assignments}
\end{figure}

\begin{figure}[h!]
\centering
\begin{subfigure}{.5\textwidth}
    \centering
    \includegraphics[width=1.0\textwidth]{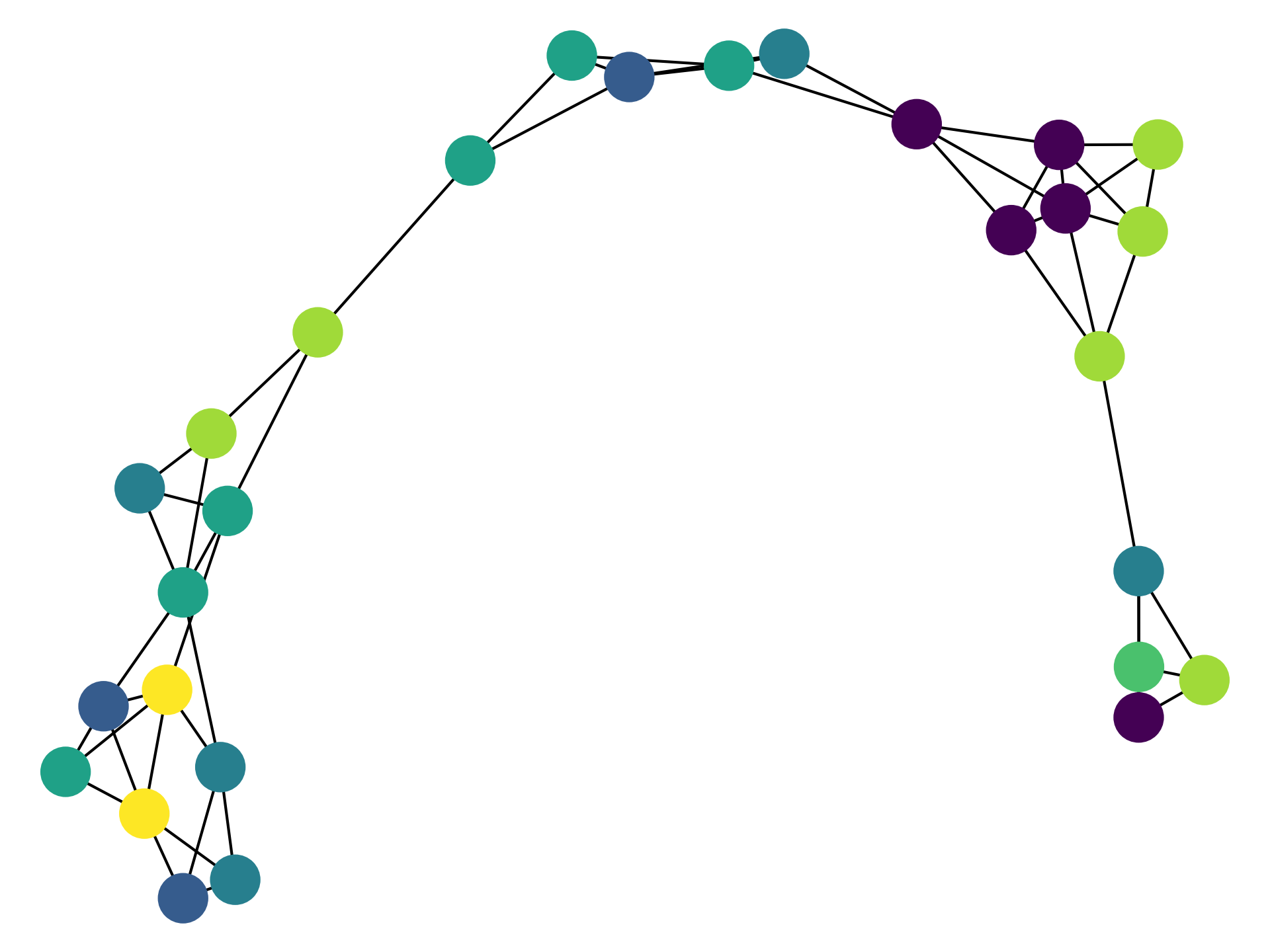}
    \caption{}
\end{subfigure}%
\begin{subfigure}{.5\textwidth}
    \centering
    \includegraphics[width=1.0\textwidth]{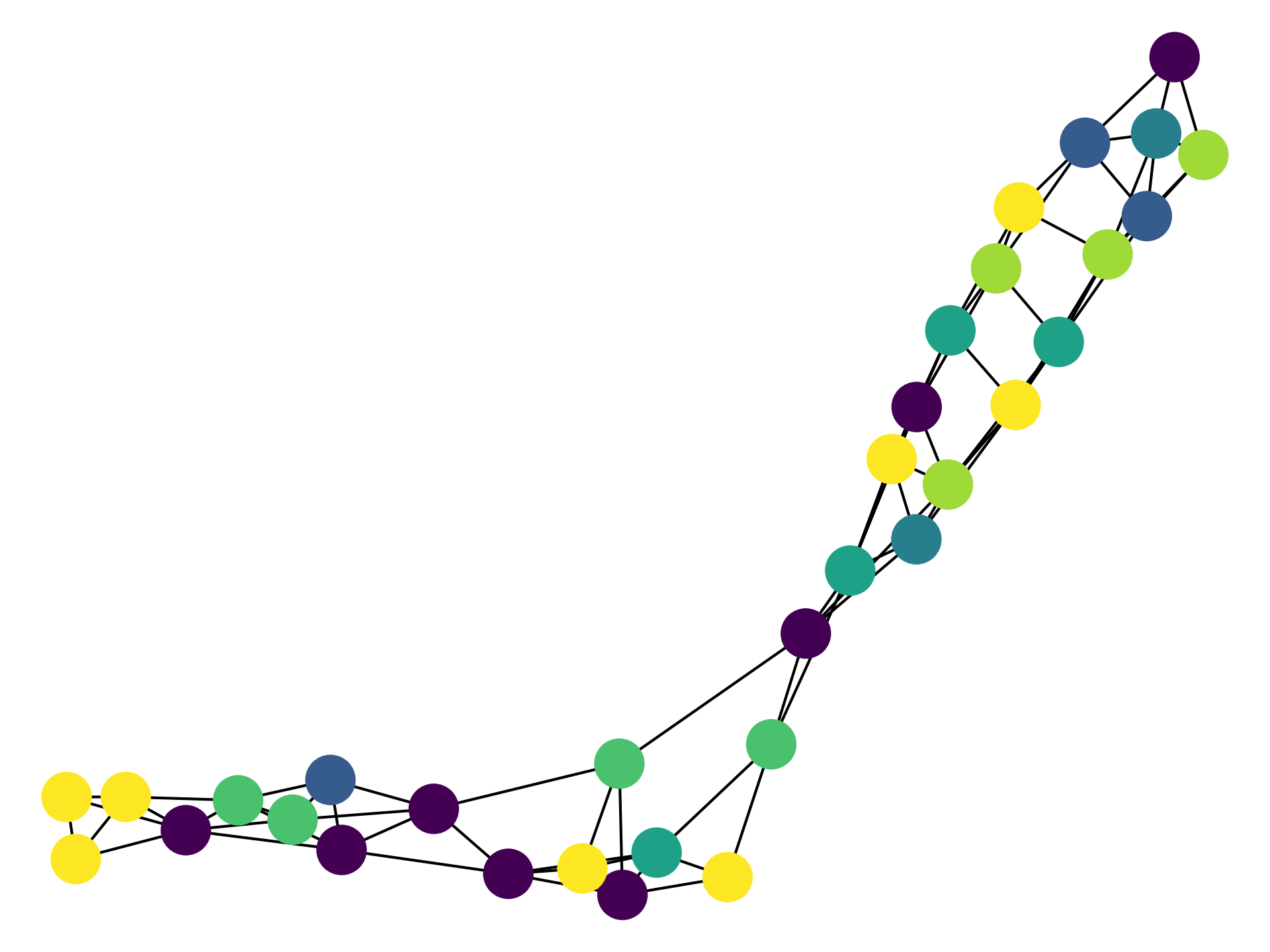}
    \caption{}
\end{subfigure}
\begin{subfigure}{.5\textwidth}
    \centering
    \includegraphics[width=1.0\textwidth]{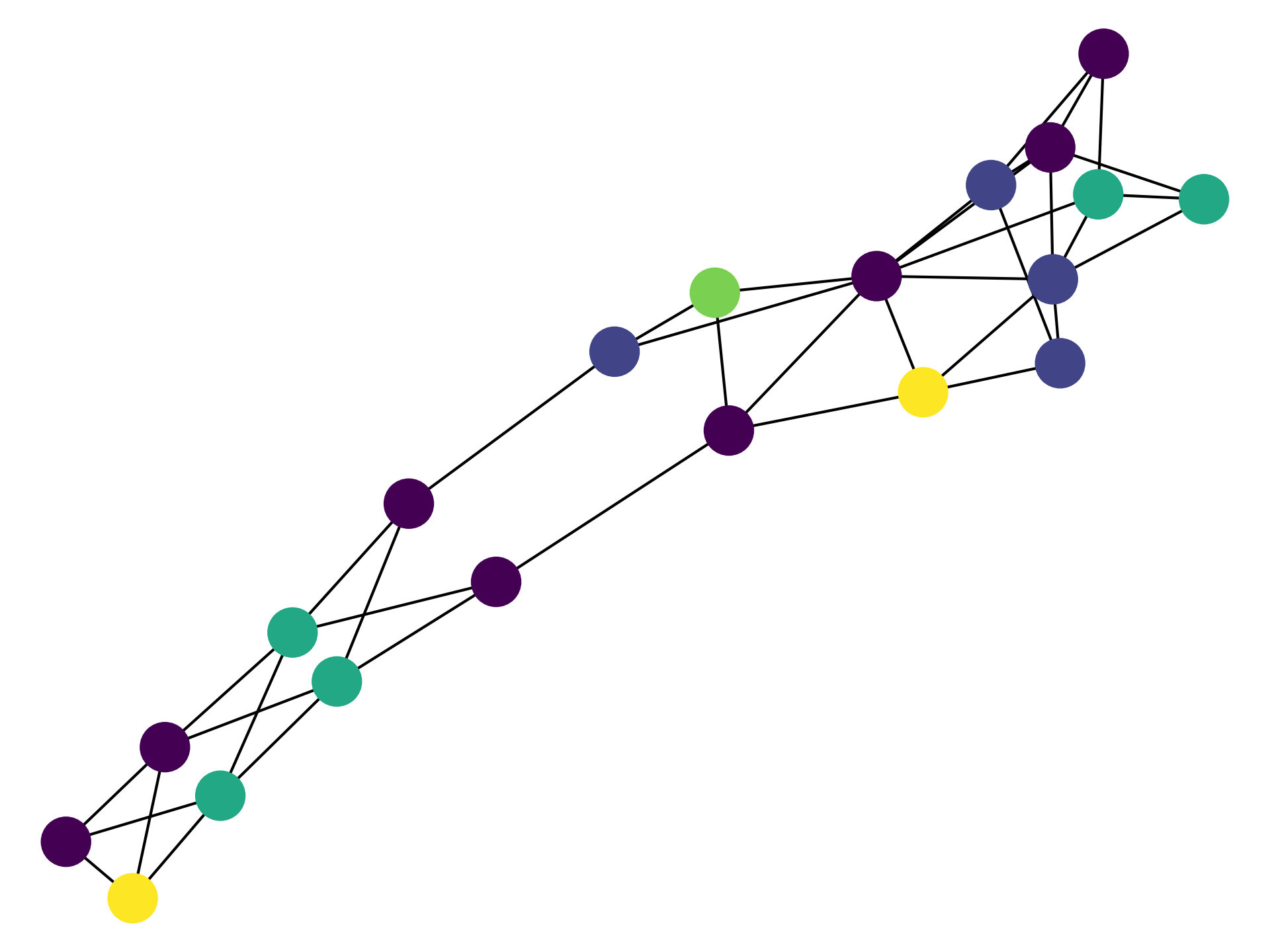}
    \caption{}
\end{subfigure}%
\begin{subfigure}{.5\textwidth}
    \centering
    \includegraphics[width=1.0\textwidth]{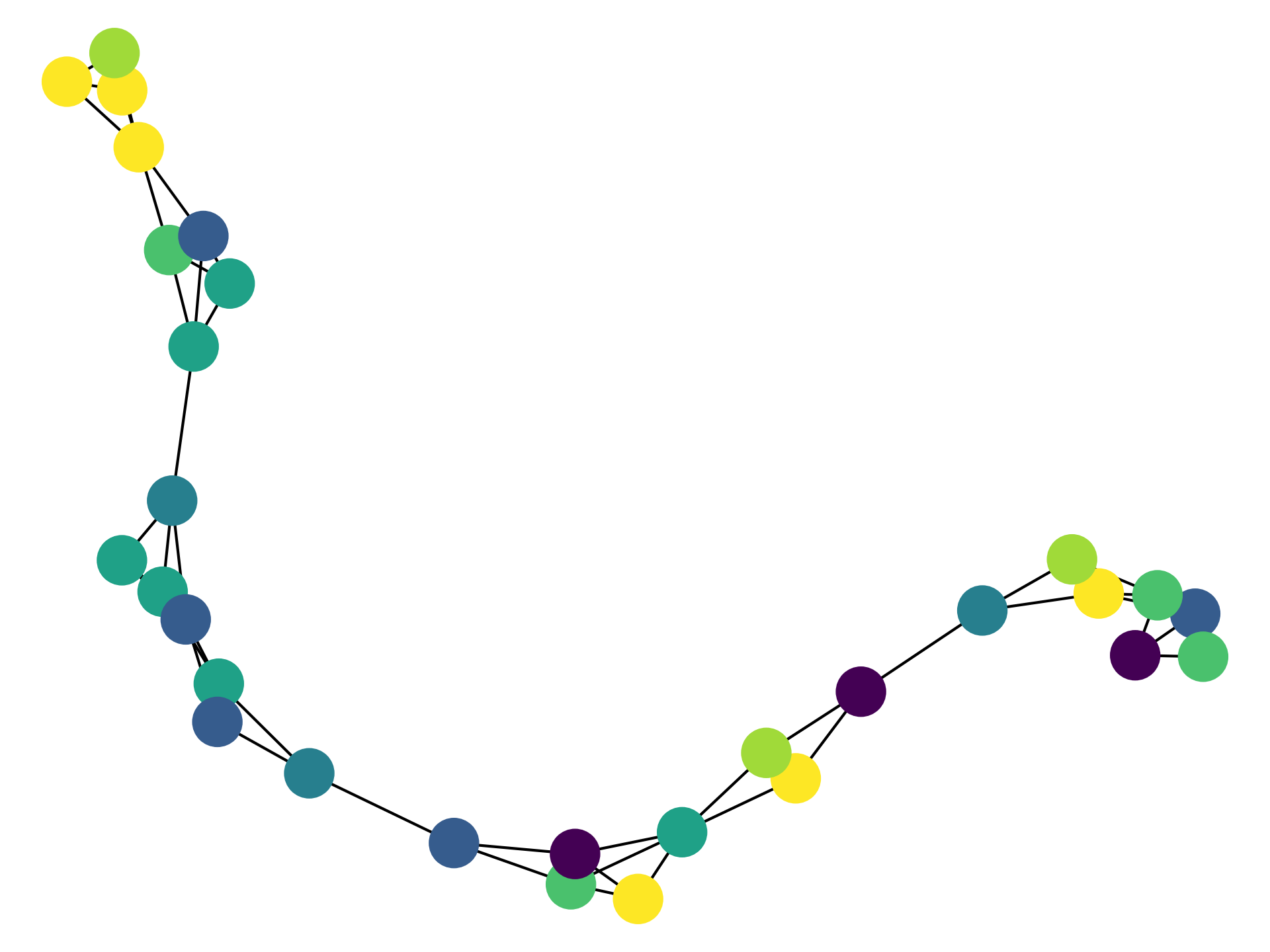}
    \caption{}
\end{subfigure}
\begin{subfigure}{.5\textwidth}
    \centering
    \includegraphics[width=1.0\textwidth]{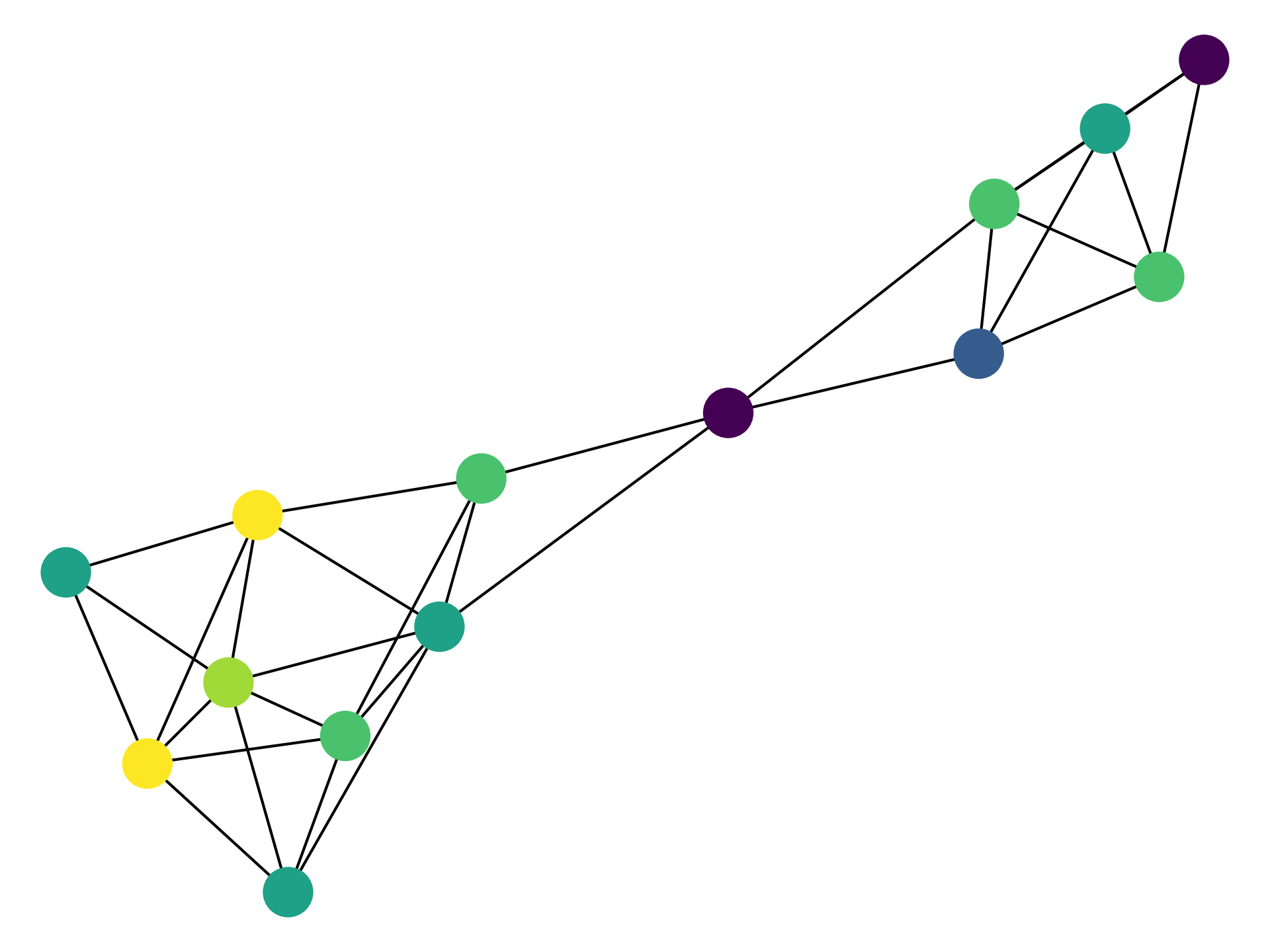}
    \caption{}
\end{subfigure}%
\begin{subfigure}{.5\textwidth}
    \centering
    \includegraphics[width=1.0\textwidth]{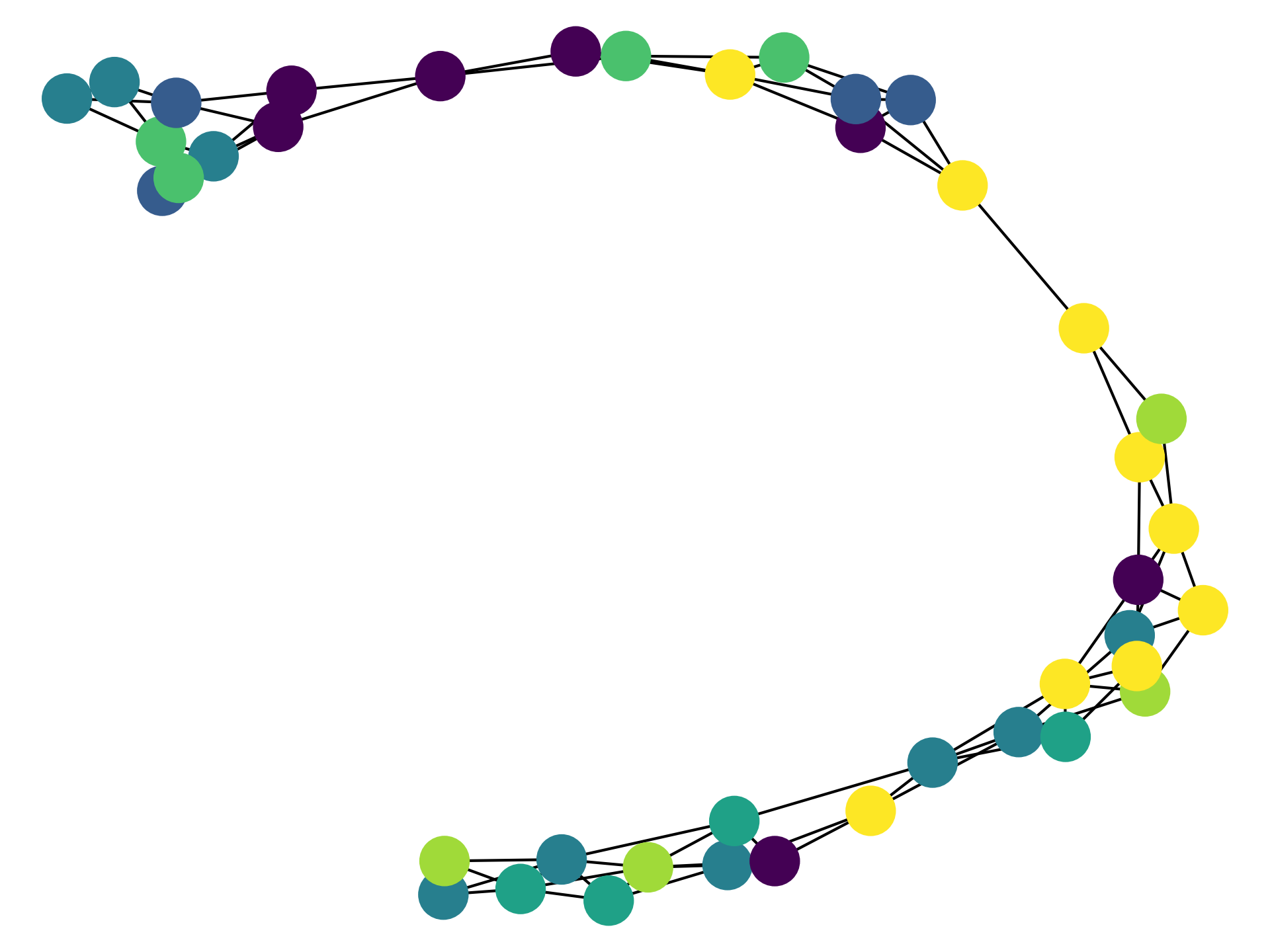}
    \caption{}
\end{subfigure}
\begin{subfigure}{.5\textwidth}
    \centering
    \includegraphics[width=1.0\textwidth]{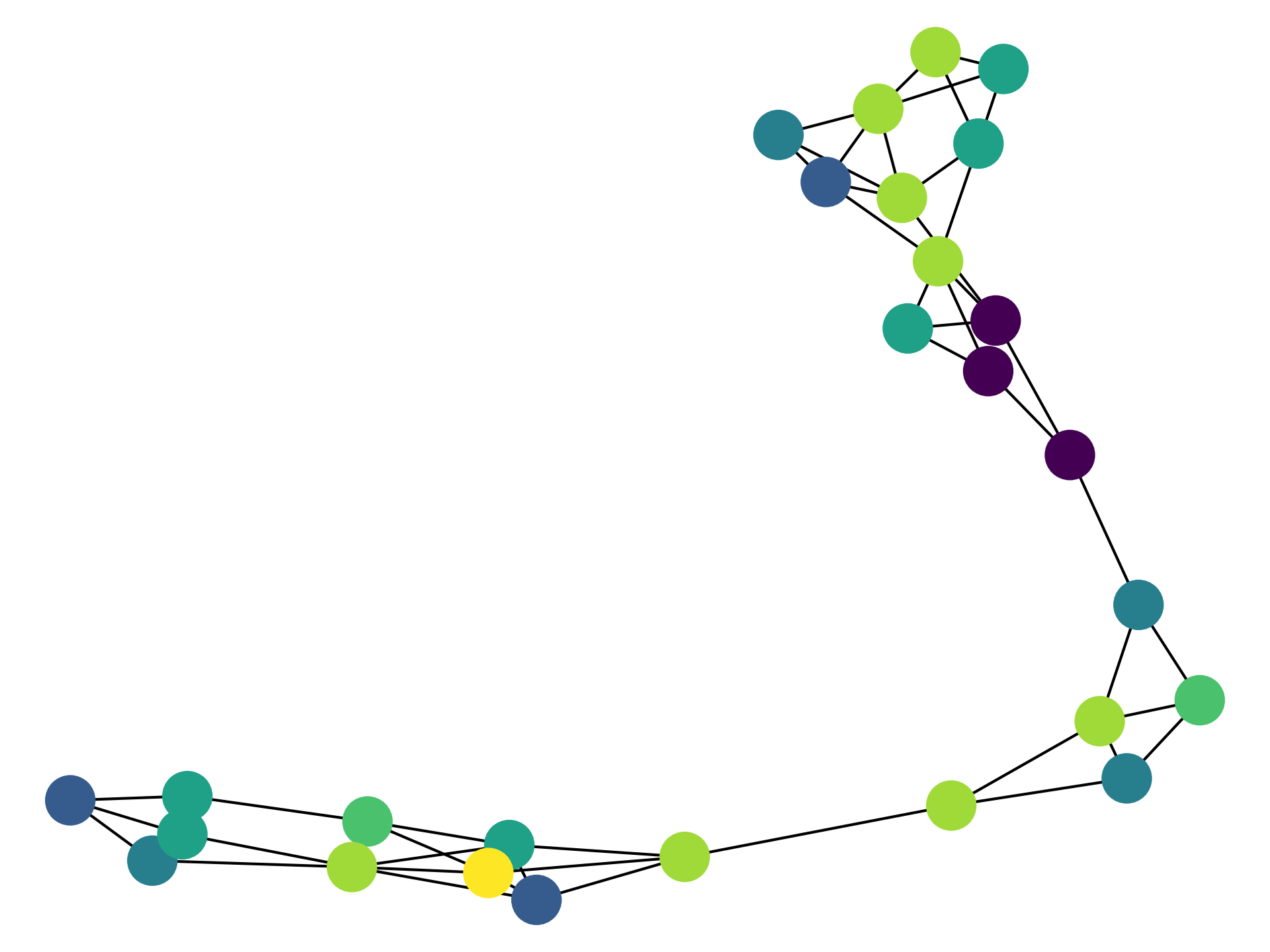}
    \caption{}
\end{subfigure}%
\begin{subfigure}{.5\textwidth}
    \centering
    \includegraphics[width=1.0\textwidth]{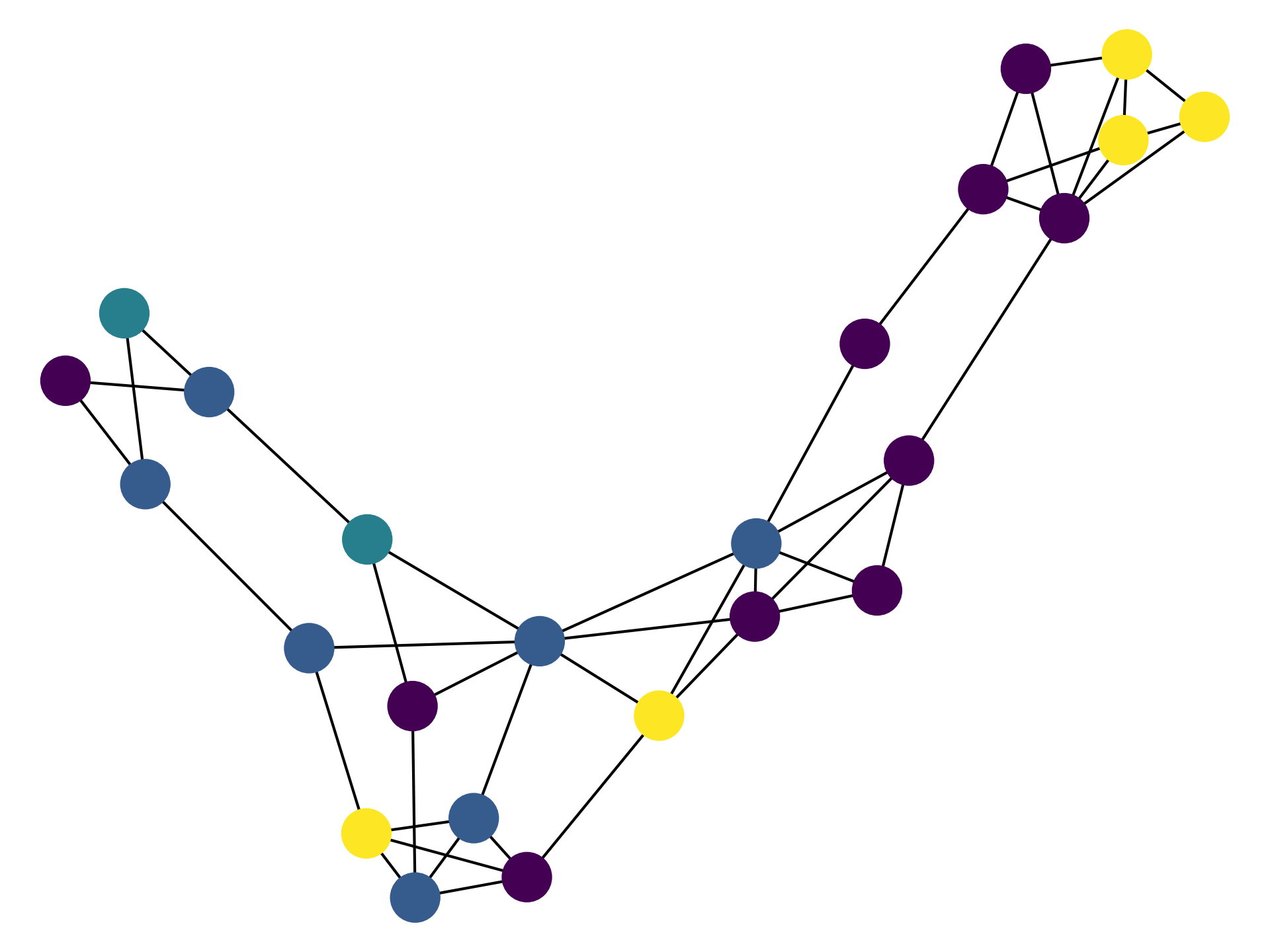}
    \caption{}
\end{subfigure}
\caption[short]{Examples of node clustering calculated by the first pooling layer using the node features for graphs in the Enzymes dataset. Node colors represent cluster assignments determined by the $argmax$ of the node assignment distribution.}
\label{fig:node cluster assignments}
\end{figure}

Figure \ref{fig:ssf cluster assignments} illustrate cluster assignment for a number of random graphs chosen from the Enzymes dataset using the proposed structural similarity features combined with the indices mapping trick. In comparison figure \ref{fig:node cluster assignments} illustrates cluster assignments by the same model using the node features for cluster assignments calculations. The plots indicate that the proposed algorithm achieves its stated goals and clusters the nodes in a manner that generally preserves localisation of nodes in the coarsened graph whereas using the node features does not generally preserve topology and is hard to interpret in terms of graph structure. In addition the number of nodes in each cluster in figure \ref{fig:ssf cluster assignments} seems to be reasonably uniform yet another indication the proposed algorithm is successful in achieving its stated goals.

\begin{table}[t]
\centering
\begin{tabular}{l c c c c c c c} 
\toprule
\multirow{2}{*}{Dataset} & \multicolumn{3}{c}{Max. Validation Accuracy} & \multicolumn{1}{c}{ } & \multicolumn{3}{c}{Epoch} \\ [0.5ex] \cline{2-4} \cline{6-8}
\multicolumn{1}{l}{} & \multicolumn{1}{c}{Structural} & \multicolumn{1}{c}{Both} & \multicolumn{1}{c}{Node} & \multicolumn{1}{c}{ } & \multicolumn{1}{c}{Structural} & \multicolumn{1}{c}{Both} & \multicolumn{1}{c}{Node} \\ [0.5ex] 
\midrule
Enzymes & \textbf{0.7667} & 0.7 & 0.7167 & \multicolumn{1}{c}{ } & 53 & 80 & 33  \\ 
D\&D & \textbf{0.7915} & 0.7881 & 0.7712 & \multicolumn{1}{c}{ } & 35 & 48 & 140 \\ 
\bottomrule
\end{tabular}
\caption{\label{tab:max accuracy}Maximal accuracy obtained on the validation data for the experiments in section \ref{S:Experimental Results:Assignment features}. The Epoch columns denote the first epoch where the maximal accuracy was obtained.}
\end{table}

\subsection{Top-$k$}
\label{S:Experimental Results:Top-k}
In this section the effect of application of the indices mapping trick \eqref{eq:Indices Mapping Trick} is evaluated by repeating the task with diminishing values of $k$ to simulate increasing information loss. In addition an experiment is conducted for evaluating the structural similarity features with no information loss and permutation invariance by modifying the $S^{(0)}$ GMN encoder such that $MLP_{node}$ is replaced with a $GRU_{node}$ core whilst maintaining the same activation and number of units. The results indicate that the choice of $k$ has substantial impact on the maximum accuracy obtained and that increasing $k$ beyond a certain point does note necessarily improve performance. Furthermore the indices mapping trick seems to substantially increase performance and using less parameters and computational resources.

\begin{table}[t]
\centering
\begin{tabular}{c c c c c c c}
\toprule
\multicolumn{7}{c}{Max. Validation Accuracy} \\ [0.5ex]
\multicolumn{1}{c}{$k$=3} & \multicolumn{1}{c}{$p$=6} & \multicolumn{1}{c}{$p$=9} & \multicolumn{1}{c}{$k$=12} & \multicolumn{1}{c}{$k$=15} & \multicolumn{1}{c}{$k$=18} & \multicolumn{1}{c}{RNN and $C^{cos,p}$} \\ [0.5ex] 
\midrule
\multicolumn{1}{c}{0.6833} & \multicolumn{1}{c}{0.6833} & \multicolumn{1}{c}{0.7} & \multicolumn{1}{c}{0.7667} & \multicolumn{1}{c}{$\mathbf{0.8}$} & \multicolumn{1}{c}{0.7167} & \multicolumn{1}{c}{0.733} \\ 

\bottomrule
\end{tabular}
\caption{\label{tab:top k}Maximal accuracy obtained on the validation data of the Enzymes dataset for different values of $k$ as well as the use of the structural similarity matrix in its entirety with an RNN core.}
\end{table}

\subsection{Similarity neighbourhood size}
\label{S:Experimental Results:Similarity neighbourhood size}
In this section the effect of modifying the neighbourhood size $p$ in equations \eqref{eq:Symmetric C} and \eqref{eq:Asymmetric C} is evaluated. For this purpose only the "structural" variant was used. The validation accuracy results are summarised in table \ref{tab:neighbourhood size}. The results indicate that a smaller neighbourhood size yielded better accuracy by a small margin in the tested settings. Other aspects of the training such as the loss values and number of clusters used by the model were observed to be similar across the different settings of $p$ used in this experiment. These results may be explained due to SimPool retaining information from all nodes across pooled layers and therefore there are little or no benefits in increasing artificially the graph connectivity as expressed in the adjacency matrix.

\begin{table}[t]
\centering
\begin{tabular}{l c c c c c c c} 
\toprule
\multirow{2}{*}{Dataset} & \multicolumn{3}{c}{Max. Validation Accuracy} & \multicolumn{1}{c}{ } & \multicolumn{3}{c}{Epoch} \\ [0.5ex] \cline{2-4} \cline{6-8} 
\multicolumn{1}{l}{} & \multicolumn{1}{c}{$p$=1} & \multicolumn{1}{c}{$p$=2} & \multicolumn{1}{c}{$p$=3} & \multicolumn{1}{c}{ } & \multicolumn{1}{c}{$p$=1} & \multicolumn{1}{c}{$p$=2} & \multicolumn{1}{c}{$p$=3} \\ [0.5ex] 
\midrule
Enzymes & \textbf{0.7667} & 0.7 & 0.7 & \multicolumn{1}{c}{ } & 53 & 78 & 32  \\ 
D\&D & \textbf{0.7915} & 0.7881 & 0.7797 & \multicolumn{1}{c}{ } & 35 & 113 & 162 \\ 
\bottomrule
\end{tabular}
\caption{\label{tab:neighbourhood size}Maximal accuracy obtained on the validation data for different neighbourhood sizes used for calculating the structural similarity features. The Epoch columns denote the first epoch where the maximal accuracy was obtained.}
\end{table}

\subsection{Regularisation term $L_C$}
\label{S:Experimental Results:Regularisation term L_C}
\begin{figure}[h!]
\centering
\begin{subfigure}{.5\textwidth}
    \centering
    \includegraphics[width=1.0\textwidth]{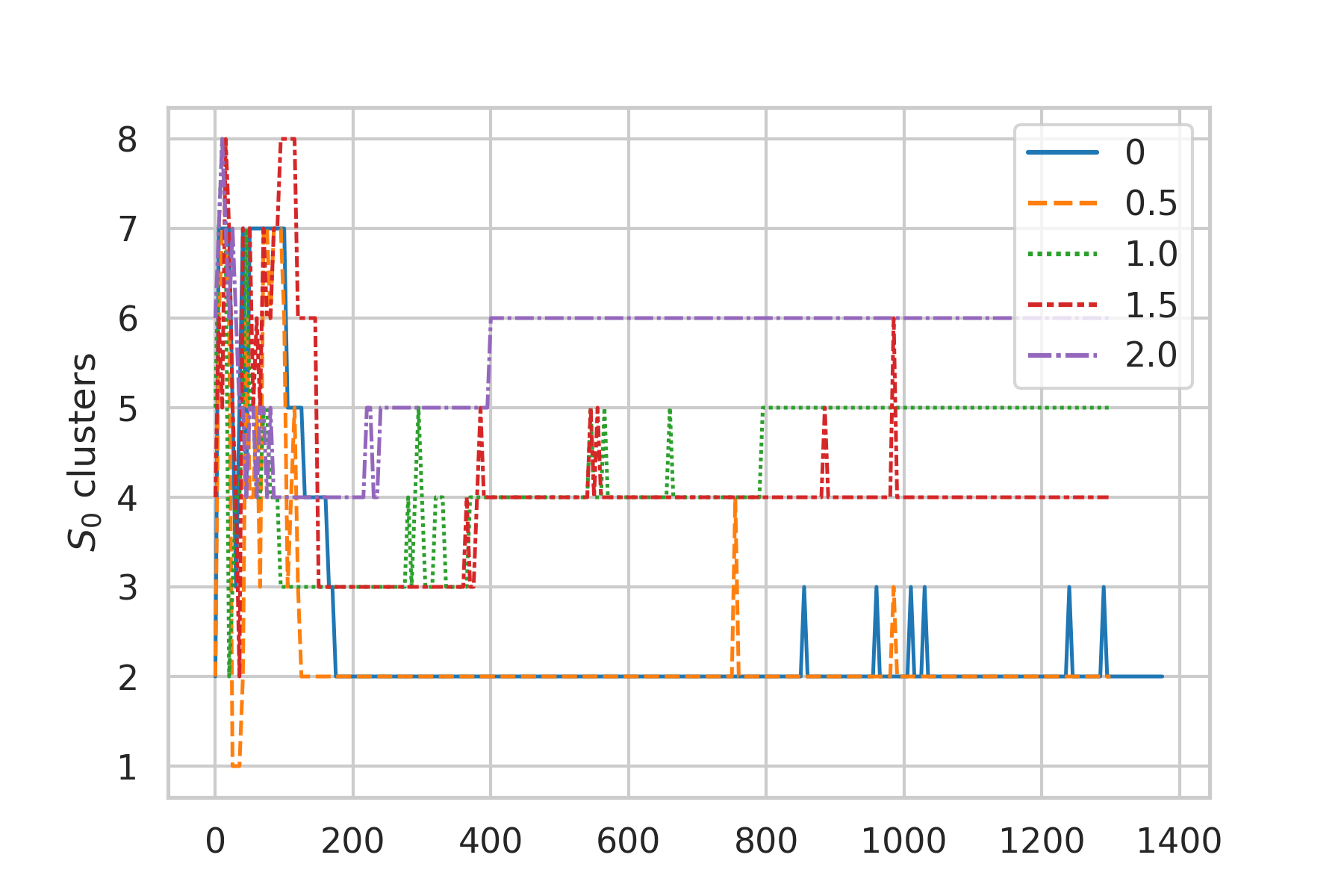}
    \caption{}
\end{subfigure}%
\begin{subfigure}{.5\textwidth}
    \centering
    \includegraphics[width=1.0\textwidth]{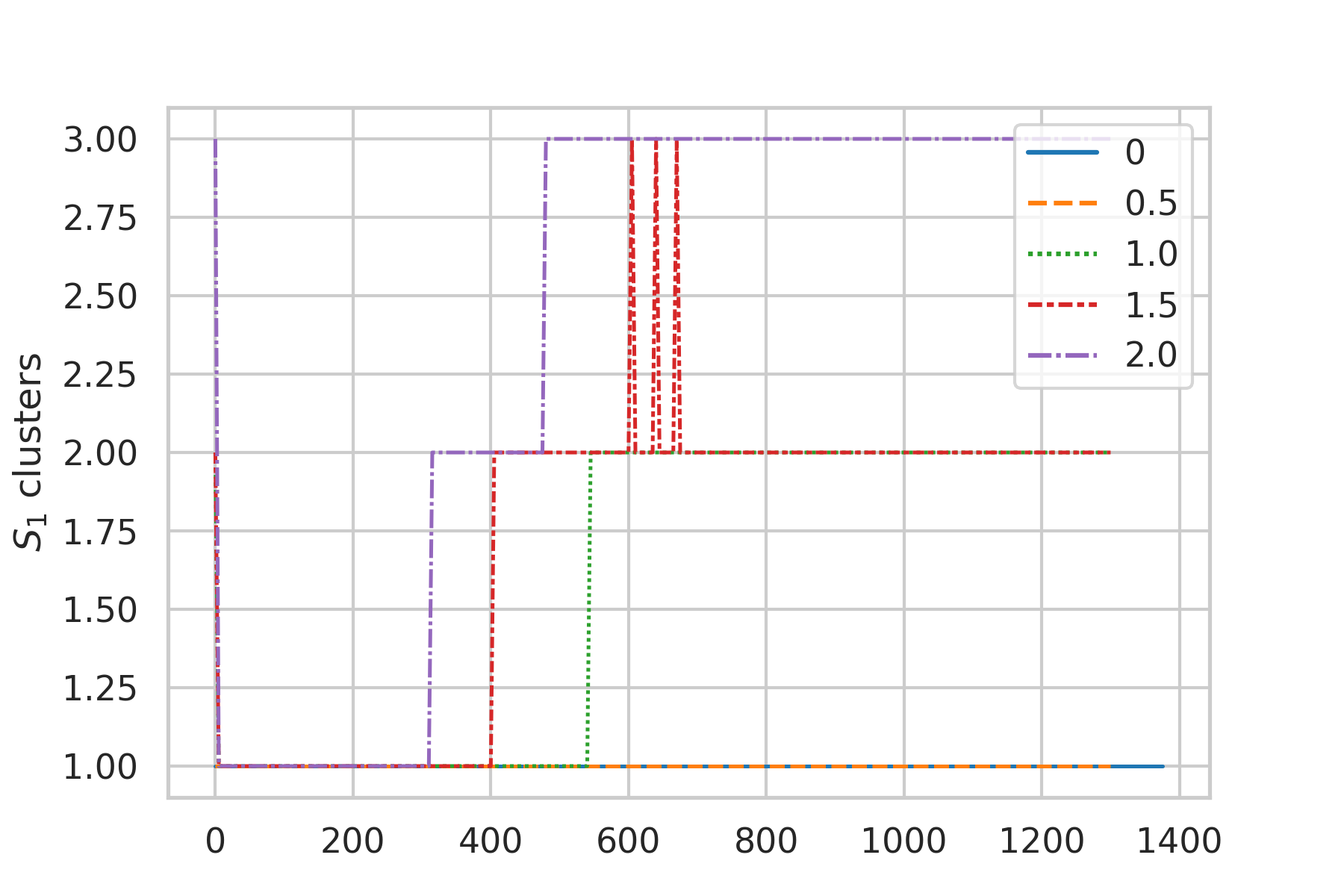}
    \caption{}
\end{subfigure}
\begin{subfigure}{.5\textwidth}
    \centering
    \includegraphics[width=1.0\textwidth]{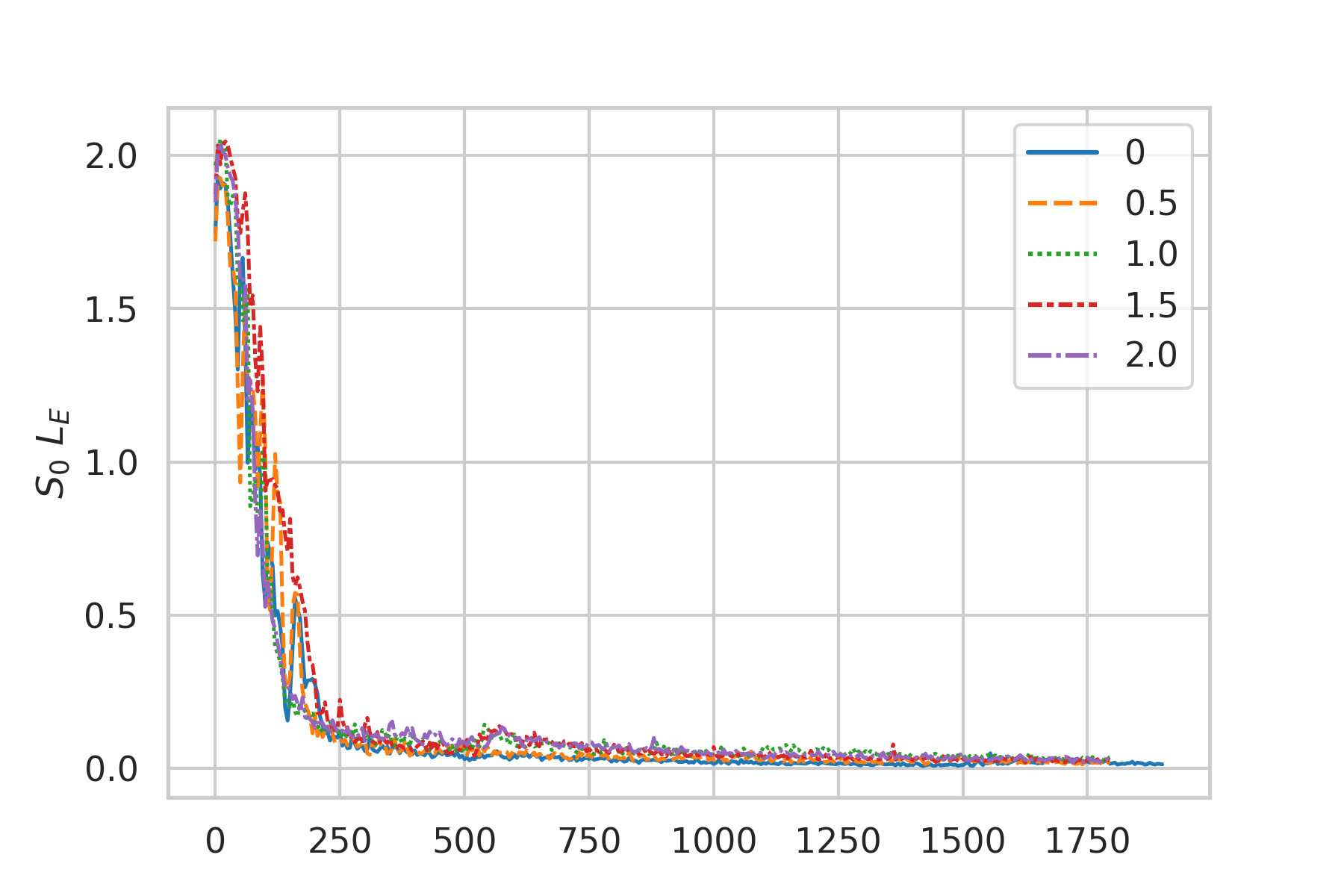}
    \caption{}
\end{subfigure}%
\begin{subfigure}{.5\textwidth}
    \centering
    \includegraphics[width=1.0\textwidth]{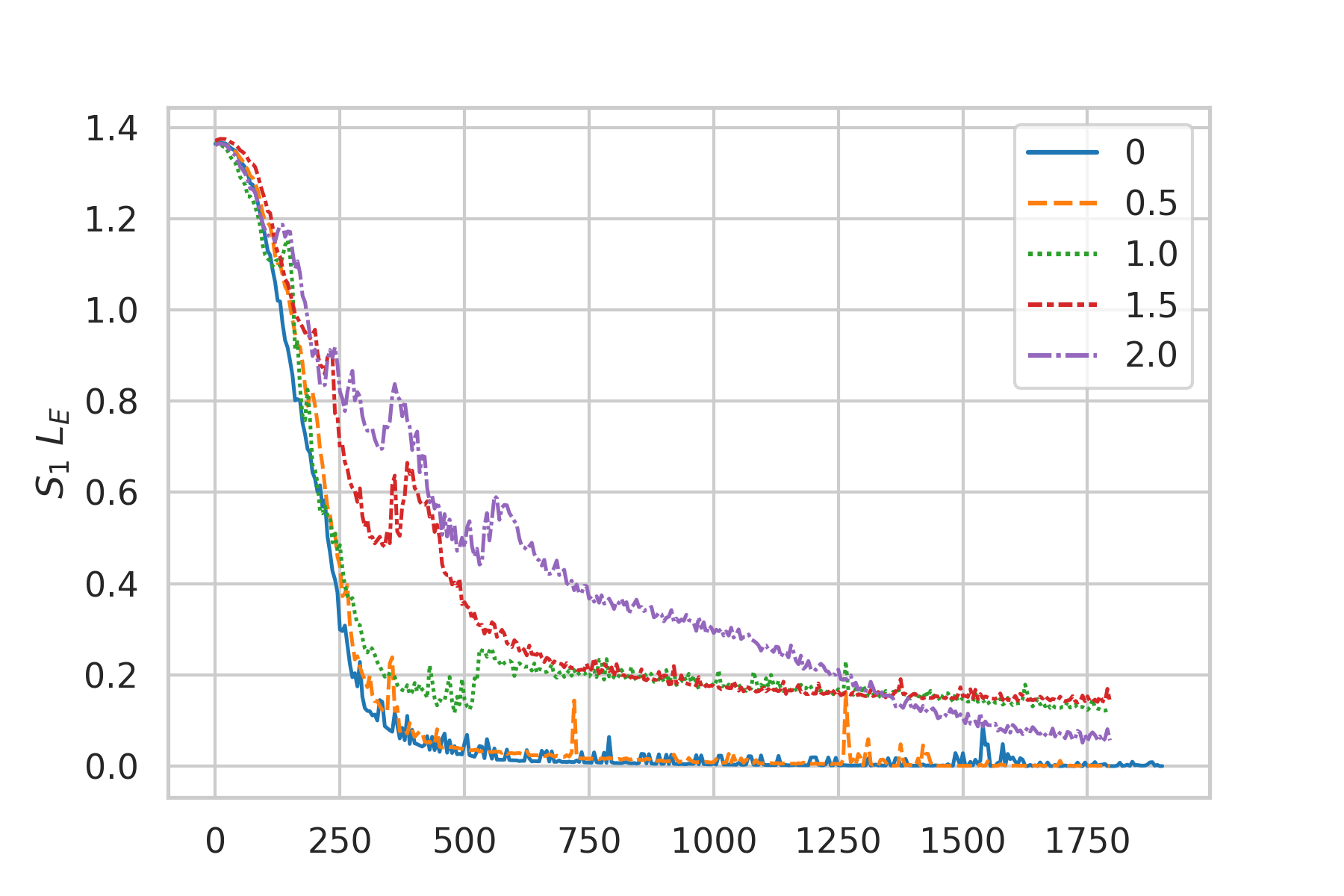}
    \caption{}
\end{subfigure}
\caption[short]{Training statistics of the Enzymes dataset for both assignment matrices with different scalers of the $L_C$ loss. $S^{(0)}$ (left column) and $S^{(1)}$ (right column). (a) and (b) number of different clusters selected; (c) and (d) loss term $L_E$ representing the uniqueness of nodes cluster assignments.}
\label{fig:LC num clusters}
\end{figure}

In this section the impact of increasing the $L_C$ loss term \eqref{eq:L_C} is explored. Figure \ref{fig:LC num clusters} illustrate values of the loss term $L_E$ during training and the number of different clusters utilised where a node cluster assignment is determined by the $argmax$ of the corresponding row in the assignment matrix.  It is evident that increasing $L_C$ encourages the model to increase the number of assigned clusters in the subsequent pooling layer possibly in exchange for increasing cluster assignment softness. However it is notable that in the first pooling layer the model was able to learn assignment policies that utilise increasing number of clusters while maintaining a close to unique assignment of nodes.

\subsection{Performance study}
\label{S:Experimental Results:Performance study}
To complete the experimental analysis a number of inductive graph classification tasks are performed and the obtained accuracy is compared against recent similar methods. For this experiment full 10-fold cross validation is performed in all experiments. The overall architecture, units and activations are summarised in table \ref{tab:architecture}, the parameters used by the SimPool layers are summarised in table \ref{tab:accuracy compare params} and table \ref{tab:accuracy compare} summarises the accuracy results obtained. To calculate the results the maximum accuracy achieved in each fold is used and their mean and standard deviation is calculated.  The results stated for methods other than SimPool are taken from the referenced papers. Since there are differences in evaluation methods across different sources the results are indicative and should not be considered as accurate benchmarking.

\begin{table}[t]
\centering
\begin{tabular}{l c c c c c c c}  
\toprule
Dataset & $p$ & $k^{(0)}$ & $\alpha^{(0)}$ & $\lambda^{(0)}$ & $L_E$ scaler & $L_C$ scaler & epochs \\ [0.5ex] 
\midrule
Enzymes &  \multirow{2}{*}{1} & 15 & \multirow{2}{*}{0.8} & \multirow{2}{*}{0} & 1.0 & \multirow{2}{*}{1.0} & 100 \\ 
D\&D & & 32 & & & 0.4 & & 60\\ 
\bottomrule
\end{tabular}
\caption{\label{tab:accuracy compare params}Parameter values used for all experiments in section \ref{S:Experimental Results:Performance study} by dataset. These values may not be optimal.}
\end{table}

\begin{table}[t]
\centering
\begin{tabular}{l  c c c}  
\toprule
\multirow{2}{*}{Method} & \multicolumn{2}{c}{Dataset} \\[0.5ex] \cline{2-3}
\multicolumn{1}{l}{} & \multicolumn{1}{c}{Enzymes} & \multicolumn{1}{c}{D\&D} \\
\midrule
\multicolumn{1}{l}{DiffPool \cite{NIPS2018_7729}} & \multicolumn{1}{c}{0.6253} & \multicolumn{1}{c}{0.8064} \\
\multicolumn{1}{l}{SortPool \cite{Zhang2018AnED}} & \multicolumn{1}{c}{0.5712 \cite{NIPS2018_7729}} & \multicolumn{1}{c}{$0.7937 \pm 0.0094$ \cite{Zhang2018AnED}} \\
\multicolumn{1}{l}{g-U-Nets \cite{DBLP:conf/icml/GaoJ19}} & \multicolumn{1}{c}{-} & \multicolumn{1}{c}{$0.8243$} \\
\multicolumn{1}{l}{SAGPool$_{h}$ \cite{35137c0f4e904fc0ab786021ead07852}} & \multicolumn{1}{c}{-} & \multicolumn{1}{c}{$0.7645 \pm 0.0097$} \\
\multicolumn{1}{l}{\textbf{SimPool}} & \multicolumn{1}{c}{$\mathbf{0.7244} \pm 0.0485$} & \multicolumn{1}{c}{TBD} \\

\bottomrule
\end{tabular}
\caption{\label{tab:accuracy compare}Comparison of accuracy achieved for different methods and datasets. The results stated for methods other than SimPool are taken from the referenced papers. Since there are differences in evaluation methods across different sources the results are indicative and should not be considered as accurate benchmarking.}
\end{table}

\begin{table}[t]
\centering
\begin{tabular}{l c c c c c} 
\toprule
\multicolumn{1}{l}{} & \multicolumn{5}{c}{$L_C$ Scaler} \\ [0.5ex] \cline{2-6}
\multicolumn{1}{l}{} & \multicolumn{1}{c}{0} & \multicolumn{1}{c}{0.5} & \multicolumn{1}{c}{1.0} & \multicolumn{1}{c}{1.5} & \multicolumn{1}{c}{2.0} \\ [0.5ex]
\midrule
\multicolumn{1}{c}{Accuracy} & \multicolumn{1}{c}{0.6833} & \multicolumn{1}{c}{0.7} & \multicolumn{1}{c}{$\mathbf{0.7667}$} & \multicolumn{1}{c}{0.7167} & \multicolumn{1}{c}{0.7} \\ 

\bottomrule
\end{tabular}
\caption{\label{tab:L_C}Maximal accuracy obtained on the validation data of the Enzymes dataset for different scalers of $L_C$.}
\end{table}

\section{Conclusion}
In this paper a method for generating features based on structural similarity that are useful for hierarchical coarsening of graphs is proposed. The method is differentiable and can integrate with many algorithms including end-to-end deep representation learning models, and can also be augmented with additional features such as the node features themselves. Furthermore SimPool is proposed, a pooling layer based on a revisited DiffPool layer that enables end-to-end GNN models to pool neighbouring nodes together in the coarsened graph encouraging the model to learn useful locality preserving pooling in a manner that is closer to the pooling operations used by CNNs that operate on local receptive fields in the standard grid. Experimental results indicate the method is successful in fulfilling its stated goals and contributes towards achieving state-of-the-art results in inductive graph classification tasks when integrated as part of an end-to-end GNN architecture.

\bibliographystyle{abbrv}
\bibliography{refs}

\end{document}